%% file: arxiv.tex
\documentclass[conference]{IEEEtran}
\usepackage{times}
\usepackage{graphicx}
\usepackage{amsmath}
\usepackage[bookmarks=true]{hyperref}
\usepackage[numbers, sort]{natbib}
\usepackage{multicol}
\usepackage{color-edits}
\usepackage[dvipsnames]{xcolor}
\usepackage{colortbl}
\usepackage{caption}
\usepackage{comment}
\usepackage{booktabs}
\usepackage{multirow}
\usepackage{url}
\usepackage{subcaption}
\usepackage{pdfpages}
\usepackage{booktabs}
\usepackage{balance}
\usepackage{tabularx}
\usepackage{adjustbox}
\usepackage{comment}
\usepackage{fancyhdr} 
\usepackage{xspace} 
\usepackage{wrapfig}
\usepackage[many]{tcolorbox}
\usepackage{amsfonts} 
\usepackage{cleveref}
\usepackage{algorithm}
\usepackage[noend]{algpseudocode}
\usepackage{bm}

\usepackage{listings}

\newbool{showauthors}
\newbool{showcomments}

\newcommand\nnfootnote[1]{%
  \begin{NoHyper}
  \renewcommand\thefootnote{}\footnote{#1}%
  \addtocounter{footnote}{-1}%
  \end{NoHyper}
}

\setbool{showauthors}{true} 
\setbool{showcomments}{true}

\newcommand{\todo}[1]{\ifbool{showcomments}{\textcolor{red} {#1}}{}}
\newcommand{\note}[3]{\ifbool{showcomments}{\textcolor{#1}{[\textbf{#2}: #3]}}{}}
\newcommand{\oursystem}{\texttt{Human2LocoMan}\xspace}

\newcommand{\parab}[1]{\noindent\textbf{#1}}
\newcommand{\parai}[1]{\noindent\textit{#1}}

\definecolor{cmured1}{HTML}{C41230}
\definecolor{cmured2}{HTML}{941120}
\definecolor{human_data}{HTML}{3A7E22}
\definecolor{robot_data}{HTML}{8B51D2}
\definecolor{body_motion}{HTML}{FF0000}
\definecolor{eef_motion}{HTML}{00F400}
\definecolor{gripper_action}{HTML}{F647F8}
\definecolor{HIT-S}{HTML}{8eaeff}
\definecolor{HIT-L}{HTML}{3c74ff}
\definecolor{MXT-Scratch-S}{HTML}{ddb2b8}
\definecolor{MXT-Scratch-L}{HTML}{c6868b}
\definecolor{MXT-Pretrained-S}{HTML}{ac5b5d}
\definecolor{MXT-Pretrained-L}{HTML}{8f302f}
\addauthor[Yunzhe]{yz}{blue}

\begin{document}

\title{\oursystem: Learning Versatile Quadrupedal Manipulation with Human Pretraining}

\author{Author Names Omitted for Anonymous Review. Paper-ID 383.}

\author{\authorblockN{Yaru Niu$^{1*}$, Yunzhe Zhang$^{1*}$, \\Mingyang Yu$^{1}$, Changyi Lin$^{1}$, Chenhao Li$^{1}$, Yikai Wang$^{1}$, Yuxiang Yang$^{2}$, Wenhao Yu$^{2}$, \\Tingnan Zhang$^{2}$, Zhenzhen Li$^{3}$, Jonathan Francis$^{1,3}$, Bingqing Chen$^{3}$, 
Jie Tan$^{2}$, and Ding Zhao$^{1}$}
\authorblockA{
$^{1}$Carnegie Mellon University \qquad $^{2}$Google DeepMind \qquad $^{3}$Bosch Center for AI \\
\href{https://human2bots.github.io/}{\textcolor{cmured2}{https://human2bots.github.io}}
}}



%

\makeatletter
\let\@oldmaketitle\@maketitle
\renewcommand{\@maketitle}{\@oldmaketitle
\centering
  \includegraphics[width=\linewidth]{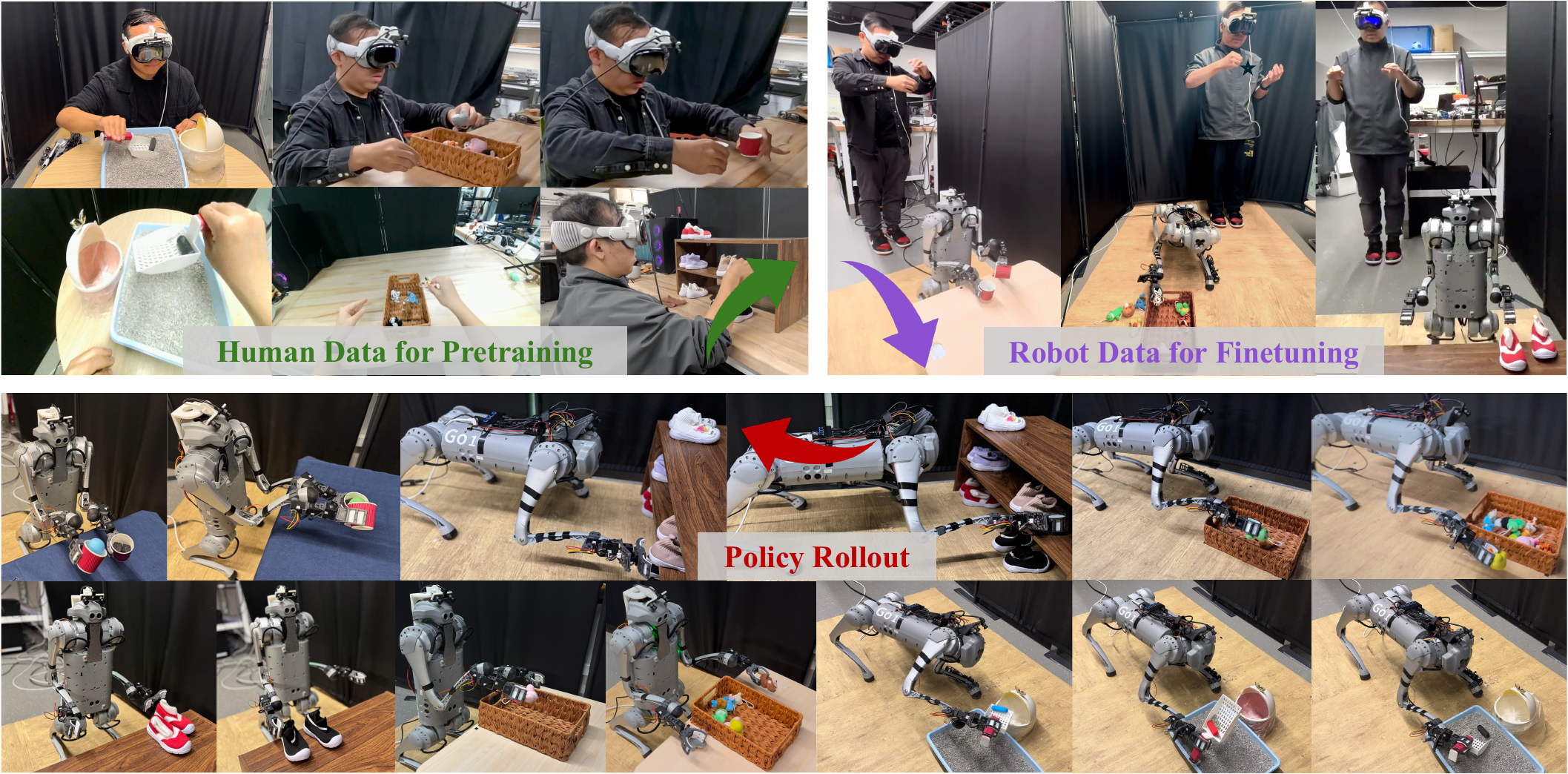}
  \captionof{figure}{
\textbf{\oursystem} provides a unified framework for collecting human demonstrations and teleoperated robot whole-body motions, along with cross-embodiment policy learning for quadrupedal manipulation. \textcolor{human_data}{Human data} is used for model pretraining, while \textcolor{robot_data}{robot data} is leveraged for policy finetuning. \oursystem achieves positive transfer from human to quadrupedal embodiments, facilitating versatile manipulation skills for unimanual and bimanual, non-prehensile and prehensile, precise tool-use, and long-horizon tasks.
  } %
  \label{fig:splash}
}
\makeatother

\let\oldmaketitle\maketitle
\renewcommand{\maketitle}{
  \oldmaketitle
  \addtocounter{figure}{-1}  
}

\maketitle

\ifbool{showauthors}{%
\nnfootnote{*Authors contributed equally to this work.}
}%

\begin{abstract}
Quadrupedal robots have demonstrated impressive locomotion capabilities in complex environments, but equipping them with autonomous versatile manipulation skills in a scalable way remains a significant challenge. 
In this work, we introduce a cross-embodiment imitation learning system for quadrupedal manipulation, leveraging data collected from both humans and LocoMan, a quadruped equipped with multiple manipulation modes.
Specifically, we develop a teleoperation and data collection pipeline, which unifies and modularizes the observation and action spaces of the human and the robot. To effectively leverage the collected data, we propose an efficient modularized architecture that supports co-training and pretraining on structured modality-aligned data across different embodiments. Additionally, we construct the first manipulation dataset for the LocoMan robot, covering various household tasks in both unimanual and bimanual modes, supplemented by a corresponding human dataset.
We validate our system on six real-world manipulation tasks, where it achieves an average success rate improvement of $41.9\%$ overall and $79.7\%$ under out-of-distribution (OOD) settings compared to the baseline. Pretraining with human data contributes a $38.6\%$ success rate improvement overall and $82.7\%$ under OOD settings, enabling consistently better performance with only half the amount of robot data.
Our code, hardware, and data are open-sourced at: \href{https://human2bots.github.io/}{\textcolor{cmured2}{https://human2bots.github.io}}.
\end{abstract}

\IEEEpeerreviewmaketitle

\input{arxiv_contents/introduction}
\input{arxiv_contents/related_work}
\input{arxiv_contents/methodology}

\input{arxiv_contents/experiments}

\input{arxiv_contents/limitations}
\input{arxiv_contents/conclusion}


\bibliographystyle{unsrtnat}
\bibliography{references}

\clearpage

\input{contents/appendix}

\end{document}

%% file: arxiv_contents/introduction.tex
\section{Introduction}
\label{sec:introduction}












While quadrupedal robots have demonstrated impressive locomotion capabilities in complex environments~\cite{jenelten2024dtc, choi2023learning, lee2020learning, yang2023neural, yang2024agile, kumar2021rma, lindqvist2022multimodality}, and recent advances have extended their abilities to manipulation tasks~\cite{fu2023deep, wu2024helpful, ha2024umi, seo2024legato, he2024visual, qiu2024wildlma, lin2024locoman}, enabling autonomous and versatile quadrupedal manipulation at scale remains a major challenge.
Imitation learning has long been a fundamental approach for teaching robots complex skills through demonstrations~\cite{schaal1996learning}, with the acquisition of high-quality data being critical for achieving efficient and effective learning. Prior works have explored various strategies for collecting in-domain robot data, primarily focusing on robot arms~\cite{zhu2023viola, wang2023mimicplay, zhao2023learning, wu2024gello}, humanoid robots~\cite{fu2024humanplus, cheng2024tv, he2024omnih2o}, and quadrupeds equipped with top-mounted arms~\cite{ha2024umi, seo2024legato, yang2024ace}. However, collecting egocentric manipulation data on a quadrupedal platform like LocoMan remains underexplored. To scale up data collection for imitation learning, recent works propose leveraging simulation data~\cite{bousmalis2018using, jiang2024dexmimicgen, brohan2023rt} or human data~\cite{wang2023mimicplay, wang2024dexcap, bharadhwaj2024towards, srirama2024hrp, chen2024arcap, kareer2024egomimic}. Human data, in particular, have been used to provide high-level task guidance~\cite{wang2023mimicplay, bharadhwaj2024towards}, improve visual encoders~\cite{srirama2024hrp}, simulate in-domain robot data~\cite{wang2024dexcap, chen2024arcap}, or serve as additional egocentric data by treating humans as an alternative embodiment~\cite{kareer2024egomimic,qiu2025humanoid}. However, the effectiveness of human data for manipulation tasks involving non-traditional embodiments such as quadrupeds has yet to be demonstrated.
From another perspective, effective teleoperation or human demonstrations typically require a kinematically similar control system—either biological~\cite{cheng2024tv, qiu2025humanoid, kareer2024egomimic} or mechanical~\cite{zhao2023learning, wu2024gello, shaw2024bimanual, wang2024dexcap, tao2025dexwild}—to the target robot, or a specialized end effector~\cite{chi2024universal, ha2024umi, seo2024legato} at the end of the human operator.
However, the substantial embodiment gap between humans and quadrupedal robots poses challenges to both data collection and policy transfer.

To address these challenges, and drawing inspiration from the LocoMan platform~\cite{lin2024locoman}—a quadrupedal robot equipped with two leg-mounted loco-manipulators that offers a versatile foundation for learning manipulation skills across multiple operating modes—we propose \oursystem, a unified framework for quadrupedal manipulation learning using data collected through human teleoperation and demonstrations.
Specifically, to enable scalable data collection, our system leverages an extended reality (XR) headset to capture human motions while streaming a first-person view (during human data collection) or a first-robot view (during teleoperation) to the operator. For human data collection, the operator simply wears the XR headset and performs tasks naturally. During teleoperation, in addition to mapping human hand motions to the robot’s grippers, we also map head motions to the robot’s torso, expanding the robot’s workspace and enhancing its active sensing capabilities. The resulting target poses are passed to a whole-body controller to generate coordinated robot motions. To structure the data and bridge the embodiment gap, we align motions of the human and the quadruped within a shared unified coordinate frame.

Different from the works that use egocentric human data to pretrain vision encoders \cite{nair2022r3m, srirama2024hrp}, learn interaction plan prediction \cite{bharadhwaj2024towards}, or co-train models with data from robots that share similar kinematics with humans~\cite{kareer2024egomimic, qiu2025humanoid}, we treat the human as a distinct embodiment from the target robot and leverage human data for model pretraining. Despite mapping human and robot data to a unified frame, there exist obvious gaps ranging from differences in dynamics to extra wrist cameras on the robot. To facilitate cross-embodiment learning and preserve modality-specific distributions unique to each embodiment, we design a modular Transformer architecture, \textit{\textbf{M}odularized \textbf{Cross}-embodiment \textbf{T}ransformer} (MXT), which shares a common Transformer trunk across embodiments while maintaining embodiment-specific tokenizers and detokenizers for shared modalities. The MXT policy is first pretrained on human data and subsequently finetuned with a small amount of robot data. A single pretrained model can be adapted to different robot embodiments through finetuning. Notably, our approach is orthogonal to prior work that leverages egocentric human videos to pretrain visual encoders or co-trains models with target robot data. Our architecture is compatible with any pretrained visual encoder and supports co-training with multi-embodiment data during both the pretraining and finetuning stages. We evaluate our approach on six household tasks across both unimanual and bimanual manipulation modes, achieving a $41.9\%$ average improvement and $79.7\%$ under OOD settings over the baseline. We also find that pretraining with human data boosts success by $38.6\%$ overall and $82.7\%$ under OOD scenarios, demonstrating effective positive transfer from humans to quadrupedal embodiments despite the large embodiment gap. These results highlight the effectiveness of our system for learning versatile quadrupedal manipulation skills and underscore its potential for scalable, large-scale cross-embodiment learning.


In summary, our paper provides the following contributions:
\begin{itemize}
    \item We propose \oursystem, a framework that enables flexible and scalable collection of human demonstrations and teleoperated robot trajectories for learning versatile quadrupedal manipulation skills.
    \item We design MXT, a modularized Transformer architecture that facilitates effective cross-embodiment learning despite large embodiment gaps between humans and quadrupedal robots.
    \item We introduce the first XR-based teleoperation system and manipulation dataset for the open-source LocoMan \cite{lin2024locoman} hardware platform.
    \item We demonstrate positive human-to-robot transfer, high success rates, and strong robustness across six challenging household tasks, in both unimanual and bimanual manipulation modes.
\end{itemize}


%% file: arxiv_contents/related_work.tex
\section{Related Work}
\label{sec:related_work}

\parab{Embodiments for Diverse Loco-Manipulation Skills:} 
Learning manipulation skills on quadrupedal robots has shown promise and popularity in recent years, due to the versatility and mobility of the platforms. Many manipulator configurations and capabilities have been proposed for quadrupeds, including non-prehensile manipulation using the quadruped's legs or body (e.g., dribbling a soccer ball, pressing buttons, closing appliance doors, etc.) \cite{ji2023dribblebot, shi2021circus, cheng2023legs, zhang2024learning, he2024learning, huang2023bayrntune, jeon2023learning, stolle2024perceptive}, using a back-mounted arm for tabletop tasks \cite{qiu2024learning, fu2023deep}, or using leg-mounted manipulators for spatially-constrained (e.g., reaching toys underneath furniture) or bimanual manipulation tasks \cite{lin2024locoman}. In this work, we take inspiration from the open-source LocoMan hardware platform \cite{lin2024locoman}, with two leg-mounted manipulators, which enable the training of policies across challenging tasks and multiple operating modes. 

\parab{Learning Versatile Quadrupedal Manipulation:} 
Reinforcement learning (RL) has been used for training individual non-prehensile manipulation skills \cite{ji2023dribblebot, shi2021circus, zhang2024learning, he2024learning, kumar2023cascaded, huang2023bayrntune, jeon2023learning, stolle2024perceptive, feng2024learning, xiong2024mqe, an2024solving, nachum2019multi, ji2021reinforcement} and for training whole-body controllers to track end-effector poses for uni-manual grasping \cite{fu2023deep, pan2024roboduet, wu2024helpful, liu2024visual, zhang2024gamma, ha2024umi, arm2024pedipulate}; here, policies are trained in simulation then transferred to the real robot platform, often with high cost in training complexity and training time. To mitigate some of these issues, imitation learning (IL) allows robots to directly learn from expert demonstrations~\cite{argall2009survey, hussein2017imitation, ravichandar2020recent, schaal1996learning} and thus provides an alternative approach to efficiently acquiring more general manipulation skills \cite{memmel2025strap, khazatsky2024droid, brohan2022rt, brohan2023rt, shafiullah2023bringing}. However, collecting robot data for quadrupedal platforms remains challenging, due to their high degrees of freedom and the need for stable whole-body controllers. Prior works have trained non-prehensile quadrupedal manipulation policies by learning from demonstrations collected in simulation~\cite{he2024visual}, or grasping policies for a top-mounted arm using data collected from real-world demonstrations~\cite{ha2024umi, seo2024legato, qiu2024wildlma}. Our work introduces a scalable way of achieving more versatile manipulation skills on quadrupedal platforms encompassing both unimanual and bimanual manipulation tasks, using a small amount of robot data combined with human demonstrations collected via our novel teleoperation and data collection system.

\parab{Data Collection for Imitation Learning:}
Various methods have been utilized to collect data for imitation learning. Joysticks and spacemouses~\cite{lin2024spawnnet, liu2022robot, zhu2023viola} are commonly used to directly teleoperate the robot for data collection. Cameras are employed to capture human motions and map them to the robot~\cite{fu2024humanplus, he2024h2o, zhang2018real, wang2023mimicplay, poubel2014support}. VR controllers provide a more intuitive way for the human to teleoperate the robot with visual or haptic feedback for dexterous manipulation tasks on robot arms, quadrupeds, and humanoid robots~\cite{cheng2024tv, sen2024learning, kareer2024egomimic, qiu2024wildlma, lu2024mobile, he2024omnih2o, ding2024bunny}.
While most works above teleoperate the robot in task space, other works employ ex-skeleton or leader-follower systems to collect robot demonstrations by mapping the joint positions of the leader system to the robot~\cite{yang2024ace, fang2024airexo, zhao2023learning, wu2024gello, kareer2024egomimic}.
To ease the burdens of teleoperating real robots and to scale up data collection, recent works have achieved success by collecting human demonstrations in the wild with AR-assisted~\cite{chen2024arcap} or hand-held grippers~\cite{chi2024universal, seo2024legato}, though these are limited to a specific robot or end-effector type. Other works enable more ergonomic data collection with body-worn cameras~\cite{papagiannis2024r+, wang2024dexcap} or VR glasses~\cite{kareer2024egomimic}. 
We introduce a unified framework to collect cross-embodiment data including both robot and human demonstrations, where the teleoperation system considers the whole-body motions of the embodiments to extend its workspace and actively sense the environment. The different manipulation modes of both the robot and human are regarded as different embodiments and the collected data can be used for model pretraining.

\parab{Cross-Embodiment Learning:} Drawing from the success of foundation models in computer vision and natural language, there have been many endeavors to replicate the success in robotics by training generalist policies on large-scale data from different robot embodiments \cite{kim2024openvla, team2024octo, o2024open, chen2024mirage, doshi2024scaling, wang2024scaling, black2024pi_0}, where the heterogeneity and gaps in kinematics, vision, and proprioception have to be handled.
For example, CrossFormer processes variable observation inputs by tokenizing images and proprioceptive information, predicts variable action outputs using action readout tokens, and conditions on language instructions or goal images~\cite{doshi2024scaling}. HPT provides a reusable trunk for cross-embodiment learning and encodes observations into a fixed number of tokens, explicitly balancing image and proprioception~\cite{wang2024scaling}. In our work, we propose Modularized Cross-embodiment Transformer (MXT), which adopts a modular design for both tokenization and detokenization, and further enhances modularity by identifying fine-grained alignments of data modalities across embodiments.



While these works primarily rely on robot data, human videos and demonstrations offer potentials for scaling up cross-embodiment learning. Recent robotic foundation models~\cite{team2025gemini, intelligence2025pi_, bjorck2025gr00t} leverage large-scale Internet data, including egocentric human videos, to pretrain higher-level models that enhance reasoning and semantic understanding. However, effectively utilizing such data for robot motor policy learning remains a significant challenge.
Notably, EgoMimic~\cite{kareer2024egomimic} treats humans as another embodiment and demonstrates strong positive transfer by co-training on both human and robot data. To enable such transfer, EgoMimic narrows the kinematic gap by selecting a human-like robot, reduces the proprioceptive gap through action normalization and alignment, and mitigates the appearance gap via visual masking. Our concurrent works similarly leverage human data for co-training on target robots that share similar embodiments with humans or the human hand, and apply action normalization to bridge the embodiment gap~\cite{qiu2025humanoid, tao2025dexwild}.
In comparison, \oursystem does not require observation and action normalization to align human data with a specific embodiment. Instead, it structures data into distinct modalities during collection and explicitly accounts for distributional gaps on these modalities across embodiments during training. This design enables greater flexibility and scalability for cross-embodiment learning, allowing us to achieve positive transfer from humans to multiple quadrupedal embodiments without requiring explicit data processing for domain alignment. 

%% file: arxiv_contents/methodology.tex
\section{Methodology}
\label{sec:method}

\begin{figure*}
    \centering\includegraphics[width=1.00\linewidth]{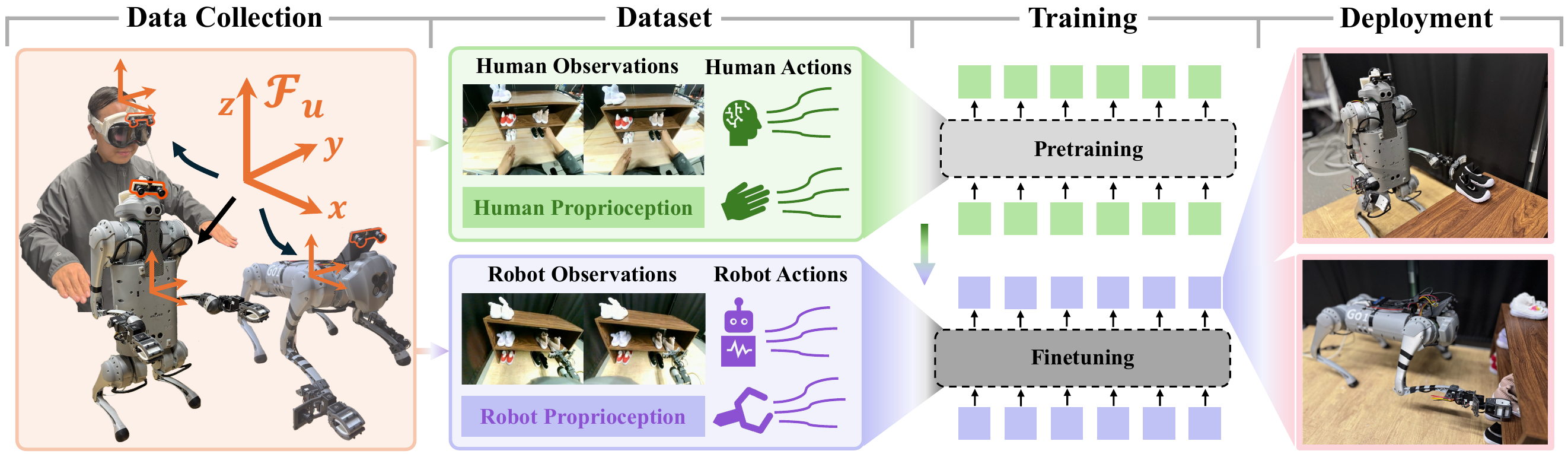}
    \caption{\textbf{\oursystem framework.} Our system uses an XR headset for data collection, capturing egocentric \textcolor{human_data}{human data} and teleoperated \textcolor{robot_data}{robot data}, all mapped to a unified coordinate frame. The dataset consists of aligned vision, proprioception, and actions from the \textcolor{human_data}{human} and the \textcolor{robot_data}{robot}. We adopt a two-stage training process: the modularized cross-embodiment model is first pretrained on easy-to-collect \textcolor{human_data}{human data}, and then finetuned on a small amount of \textcolor{robot_data}{robot data}. The resulting \oursystem policies can be deployed on real robots for versatile manipulation tasks in both unimanual and bimanual modes.}
    \label{fig:framework}
\end{figure*}


In this section, we present the design and implementation of our system, \oursystem, which integrates teleoperation, data collection, and a Transformer-based architecture for cross-embodied learning.

\subsection{\oursystem System Overview}


We utilize the Apple Vision Pro headset and the Open-Television system \cite{cheng2024tv} to capture human motions and stream first-person or first-robot video to the human operator. A lightweight stereo camera with a 120-degree horizontal field of view is mounted on both the VR headset and the LocoMan robot to provide egocentric views, while additional cameras, such as RGB wrist cameras, can be optionally attached to the robot.
Through the \oursystem teleoperation system (Section~\ref{sec:teleop}), the human operator can control the LocoMan robot to perform versatile manipulation tasks in both unimanual and bimanual modes. In the unimanual mode, we also map human head motions to the robot's torso movements to expand the teleoperation workspace and enhance active sensing.
The \oursystem system enables the collection of both human and robot data, transforming them into a shared space. Masks are applied to distinguish across different embodiments and manipulation modes. The collected human data are used to pretrain an action model called the \textit{\textbf{M}odularized \textbf{Cross}-embodiment \textbf{T}ransformer} (MXT). The in-domain robotic data collected via teleoperation  are used to  finetune the pretrained model to learn a manipulation policy that predicts the 6D poses of LocoMan’s end effectors and torso, as well as gripper actions.




\subsection{\oursystem Teleoperation and Data Collection}
\label{sec:teleop}

\parab{A unified frame for both human and LocoMan.}
To map human motions to LocoMan's various operation modes via VR-based teleoperation—and to enhance the transferability of motion data across different embodiments—we establish a unified reference frame, $\mathcal{F}_u$, to align motions across embodiments. As shown in Figure \ref{fig:framework}~(a), this unified frame is attached to the rigid body where the main camera is mounted. At the embodiment's reset pose, the x-axis points forward, aligned with the workspace and parallel to the ground; the y-axis points leftward; and the z-axis points upward, perpendicular to the ground.

\parab{Motion mapping.}
We map the human wrist motions to LocoMan's end-effector motions, map the human head motions to LocoMan's torso motions, and hand poses to LocoMan's gripper actions. The 6D poses of the human hand, head, and wrist poses in $\text{SE(3)}$ in the VR-defined world frame are streamed from the VR set to the \oursystem teleoperation server. 
The human head pose is represented as $(\bm{x}^{\text{head}}_{\text{vr}}, \bm{R}^{\text{head}}_{\text{vr}})$, and the wrist poses are $(\bm{x}^{\text{r-wrist}}_{\text{vr}}, \bm{R}^{\text{r-wrist}}_{\text{vr}})$ and $(\bm{x}^{\text{l-wrist}}_{\text{vr}}, \bm{R}^{\text{l-wrist}}_{\text{vr}})$,
where $\bm{x}^{\cdot}_{\text{vr}}$ denotes the translation and $\bm{R}^{\cdot}_{\text{vr}}$ denotes the rotation in the VR-defined world frame. Then, the 6D poses can be transformed into the unified frame $\mathcal{F}_u$ $(\bm{x}^{\cdot}_{\text{uni}}, \bm{R}^{\cdot}_{\text{uni}})=(\bm{R}^{\text{vr}}_{\text{uni}}\bm{x}^{\cdot}_{\text{vr}}
, \bm{R}^{\text{vr}}_{\text{uni}}\bm{R}^{\cdot}_{\text{vr}})$, where $\bm{R}^{\text{vr}}_{\text{uni}}$ is the rotation matrix of the VR-defined frame relative to the unified frame $\mathcal{F}_u$.

To initialize the teleoperation for each manipulation mode, the robot is transferred to a reset pose randomly initialized within a small range, termed as 
$\bm{p}_0=(\bm{x}^{\text{torso}}_{\text{uni, 0}}, \bm{R}^{\text{torso}}_{\text{uni, 0}}, \bm{x}^{\text{r-eef}}_{\text{uni, 0}}, \bm{R}^{\text{r-eef}}_{\text{uni, 0}}, \bm{x}^{\text{l-eef}}_{\text{uni, 0}}, \bm{R}^{\text{l-eef}}_{\text{uni, 0}}, \bm{\theta}^{\text{gripper}}_{0})$, including the 6D poses of the torso and both end effectors, and the gripper actions. 
The human operator starts to teleoperate the robot after an initializing posture. The target pose for the robot at time step $t$, 
$\bm{p}^{\text{t}}_t=(\bm{x}^{\text{torso,t}}_{\text{uni}, t}, \bm{R}^{\text{torso,t}}_{\text{uni}, t}, \bm{x}^{\text{r-eef,t}}_{\text{uni},t}, \bm{R}^{\text{r-eef,t}}_{\text{uni},t}, \bm{x}^{\text{l-eef,t}}_{\text{uni},t}, \bm{R}^{\text{l-eef,t}}_{\text{uni},t}, \bm{\theta}^{\text{gripper,t}}_{t})$, can be expressed as follows.
\begin{equation}
\begin{aligned}
\bm{x}^{\text{torso,t}}_{\text{uni},t} &= \bm{x}^{\text{torso}}_{\text{uni, 0}} + \alpha^{\text{torso}}( \bm{x}^{\text{head}}_{\text{uni},t}-\bm{x}^{\text{head}}_{\text{uni, 0}}) \\
\bm{R}^{\text{torso,t}}_{\text{uni},t} &= \bm{R}^{\text{torso}}_{\text{uni, 0}}((\bm{R}^{\text{head}}_{\text{uni, 0}})^\top\bm{R}^{\text{head}}_{\text{uni},t}) \\
\bm{x}^{\text{r-eef,t}}_{\text{uni},t} &= \bm{x}^{\text{r-eef}}_{\text{uni, 0}} + \alpha^{\text{r-eef}}( \bm{x}^{\text{r-wrist}}_{\text{uni},t}-\bm{x}^{\text{r-wrist}}_{\text{uni, 0}}) \\
\bm{R}^{\text{r-eef,t}}_{\text{uni},t} &= \bm{R}^{\text{r-eef}}_{\text{uni, 0}}((\bm{R}^{\text{r-wrist}}_{\text{uni, 0}})^\top\bm{R}^{\text{r-wrist}}_{\text{uni},t}) \\
\bm{x}^{\text{l-eef,t}}_{\text{uni},t} &= \bm{x}^{\text{l-eef}}_{\text{uni, 0}} + \alpha^{\text{l-eef}}( \bm{x}^{\text{l-wrist}}_{\text{uni},t}-\bm{x}^{\text{l-wrist}}_{\text{uni, 0}}) \\
\bm{R}^{\text{l-eef,t}}_{\text{uni},t} &= \bm{R}^{\text{l-eef}}_{\text{uni, 0}}((\bm{R}^{\text{l-wrist}}_{\text{uni, 0}})^\top\bm{R}^{\text{l-wrist}}_{\text{uni},t}) \\
\bm{\theta}^{\text{gripper,t}}_{t} &= \frac{\bm{\theta}^{\text{gripper}}_{\text{max}}-\bm{\theta}^{\text{gripper}}_{\text{min}}}{d^{\text{tip}}_{\text{max}}}\circ\bm{d}^{\text{tip}}_{t} + \bm{\theta}^{\text{gripper}}_{\text{min}} \\
\end{aligned}
\end{equation}
Here, $\alpha^{\text{torso}}$, $\alpha^{\text{r-eef}}$, and $\alpha^{\text{l-eef}}$, are the scaling factors to map human's motions to robot's torso, right end effector, and left end effector, respectively. $\bm{x}^{\text{gripper}}_{\text{max}}$ and $\bm{x}^{\text{gripper}}_{\text{min}}$ are the maximum and minimum gripper angles, respectively. $\bm{d}^{\text{tip}}_{t}$ represents the distances between the reference finger tips of both human hands at time step $t$, and $d^{\text{tip}}_{\text{max}}$ is the maximum finger tip distance for the human operator.

\parab{Whole-body controller.}
The robot target pose at time $t$, $p^{\text{t}}_t$, is calculated from the teleoperation server, and sent to the whole-body controller of the LocoMan robot, which is adapted from the one introduced in~\cite{lin2024locoman}, a unified whole-body controller designed to track the desired poses of the torso, end effectors, and feet across multiple operation modes. We employ null-space projection for kinematic tracking and quadratic programming for dynamic optimization to compute the desired joint positions, velocities, and torques.

To handle the large embodiment gap between the human and the LocoMan robots, and to facilitate smooth teleoperation of a dynamic quadrupedal platform with whole-body motions, we consider the handling and recovery from robot's joint limits, singularity, and self-collision, and fast motions. 
We compute the manipulability index as:
\begin{align}
    I_{\text{mani}} = \sqrt{\det(\mathbf{J} \mathbf{J}^\top)}
\end{align}
to assess the proximity of the target pose to singularity, where \( \mathbf{J} \) represents the Jacobian of the robot's manipulator at the target pose. If \( I_{\text{mani}} \) falls below a predefined threshold \( \tau_{\text{mani}} \), the target pose is considered near singularity. 
To detect self-collisions, we utilize the Pinocchio library \cite{carpentier2019pinocchio} to compute collision pairs among the robot’s body parts. If any of the following conditions are met—joint limit violation, singularity, or self-collision—the whole-body controller tracks \( p^{\text{t}}_{t-1} \) instead of \( p^{\text{t}}_t \).
To mitigate rapid movements, we apply linear interpolation between 
\( x^{\text{torso,t}}_{\text{uni},t} \) and \( x^{\text{torso,t}}_{\text{uni},t-1} \), 
\( x^{\text{r-eef,t}}_{\text{uni},t} \) and \( x^{\text{r-eef,t}}_{\text{uni},t-1} \), 
\( x^{\text{l-eef,t}}_{\text{uni},t} \) and \( x^{\text{l-eef,t}}_{\text{uni},t-1} \), 
as well as \( \theta^{\text{gripper,t}}_{t} \) and \( \theta^{\text{gripper,t}}_{t-1} \). 
Additionally, quaternion interpolation is applied between 
\( \bm{R}^{\text{torso,t}}_{\text{uni},t} \) and \( \bm{R}^{\text{torso,t}}_{\text{uni},t-1} \), 
\( \bm{R}^{\text{r-eef,t}}_{\text{uni},t} \) and \( \bm{R}^{\text{r-eef,t}}_{\text{uni},t-1} \), 
and \( \bm{R}^{\text{l-eef,t}}_{\text{uni},t} \) and \( \bm{R}^{\text{l-eef,t}}_{\text{uni},t-1} \) 
to smooth large action variations.

\begin{figure*}
    \centering
    \includegraphics[width=0.99\linewidth]{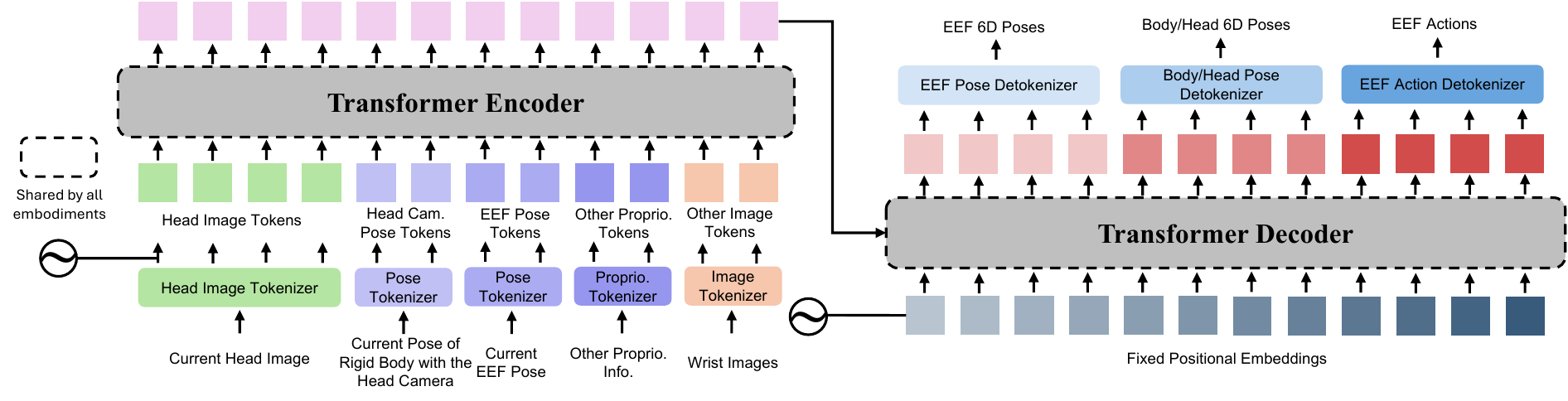}
    \caption{\textbf{Modularized Cross-embodiment Transformer (MXT) architecture.} The inputs are organized as a list of modalities and encoded each by a separate tokenizer into a fixed number of tokens. The Transformer trunk handles decision making by consuming the concatenated encoded tokens and producing a fixed number of raw output tokens. Each of the detokenizers at the end decodes a fixed subset of the output tokens into a modality of the final actions.}
    \label{fig:arch}
\end{figure*}

\parab{Data Collection.}
We record the robot data $\{\mathcal{D}^{\text{R}}_t\}^{T}_{t=1}$ during teleoperation, where $\mathcal{D}^{\text{R}}_t=\{\bm{o}^{\text{R}}_t, \bm{a}^{\text{R}}_t\}$ is the robot data at time step $t$ including the robot observations $o^{\text{R}}_t$ and robot actions $a^{\text{R}}_t$, and $T$ is the episode length. We define the $\bm{I}^{\rm R}_{{\rm main}, t}$ and $\bm{I}^{\rm R}_{{\rm wrist}, t}$ are images obtained from the robot's main stereo camera and the wrist camera, respectively. Then, we can formulate $o^{\text{R}}_t$ and $a^{\text{R}}_t$ in the dataset as follows.
\begin{equation}
\begin{aligned}
    \bm{o}^{\rm R}_t[{\rm main\ image}] &:= I_{{\rm main}, t}, \\
    \bm{o}^{\rm R}_t[{\rm wrist\ image}] &:= I_{{\rm wrist}, t}, \\
    \bm{o}^{\rm R}_t[{\rm body\ pose}] &:= [\bm{x}^{\text{torso}}_{\text{uni},t}, \bm{R}^{\text{torso}}_{\text{uni}, t}],\\
    \bm{o}^{\rm R}_t[{\rm EEF\ pose}]&:=[\bm{x}^{\text{r-eef}}_{\text{uni}, t}, \bm{R}^{\text{r-eef}}_{\text{uni}, t},\bm{x}^{\text{l-eef}}_{\text{uni}, t}, \bm{R}^{\text{l-eef}}_{\text{uni}, t}],\\
    \bm{o}^{\rm R}_t[{\rm EEF\ to\ body\ pose}]&:=[\bm{x}^{\text{r-eef}}_{\text{uni}, t} - \bm{x}^{\text{torso}}_{\text{uni}, t}, (\bm{R}^{\text{torso}}_{\text{uni}, t})^{\top}\bm{R}^{\text{r-eef}}_{\text{uni}, t} \\
    & \qquad \bm{x}^{\text{l-eef}}_{\text{uni}, t} - \bm{x}^{\text{torso}}_{\text{uni}, t}, (\bm{R}^{\text{torso}}_{\text{uni}, t})^{\top}\bm{R}^{\text{l-eef}}_{\text{uni}, t}], \\
    \bm{o}^{\rm R}_t[{\rm gripper\ actions}] &:= \bm{\theta}^{\text{gripper}}_{t}, \\
    \bm{a}^{\rm R}_t[{\rm body\ pose}] &:=[\bm{x}^{\text{torso, t}}_{\text{uni}, t}, \bm{R}^{\text{torso, t}}_{\text{uni}, t}], \\
    \bm{a}^{\rm R}_t[{\rm EEF\ pose}] &:=[\bm{x}^{\text{r-eef, t}}_{\text{uni}, t}, \bm{R}^{\text{r-eef, t}}_{\text{uni}, t},\bm{x}^{\text{l-eef, t}}_{\text{uni}, t}, \bm{R}^{\text{l-eef, t}}_{\text{uni}, t}], \\
    \bm{a}^{\rm R}_t[{\rm gripper\ actions}] &:= \bm{\theta}^{\text{gripper, t}}_{t} \label{eq:robotdata}
\end{aligned}
\end{equation}

We record the human data $\{\mathcal{D}^{\text{H}}_t\}^{T}_{t=1}$ in real time during human's manipulation. Similarly, the human data at time step $t$ $\mathcal{D}^{\text{H}}_t=\{\bm{o}^{\text{H}}_t, \bm{a}^{\text{H}}_t\}$ can be defined by human observations $\bm{o}^{\text{H}}_t$ and human actions $\bm{a}^{\text{H}}_t$ as follows.
\begin{equation}
\begin{aligned}
    \bm{o}^{\rm H}_t[{\rm main\ image}] &:= I^{\rm H}_{{\rm main}, t}, \\
    \bm{o}^{\rm H}_t[{\rm body\ pose}] &:= [\bm{x}^{\text{head}}_{\text{uni},t}, \bm{R}^{\text{head}}_{\text{uni}, t}],\\
    \bm{o}^{\rm H}_t[{\rm EEF\ pose}]&:=[\bm{x}^{\text{r-wrist}}_{\text{uni}, t}, \bm{R}^{\text{r-wrist}}_{\text{uni}, t},\bm{x}^{\text{l-wrist}}_{\text{uni}, t}, \bm{R}^{\text{l-wrist}}_{\text{uni}, t}],\\
    \bm{o}^{\rm H}_t[{\rm EEF\ to\ body\ pose}]&:=[\bm{x}^{\text{r-wrist}}_{\text{uni}, t} - \bm{x}^{\text{head}}_{\text{uni}, t}, (\bm{R}^{\text{head}}_{\text{uni}, t})^{\top}\bm{R}^{\text{r-wrist}}_{\text{uni}, t} \\
    & \qquad \bm{x}^{\text{l-wrist}}_{\text{uni}, t} - \bm{x}^{\text{head}}_{\text{uni}, t}, (\bm{R}^{\text{head}}_{\text{uni}, t})^{\top}\bm{R}^{\text{l-wrist}}_{\text{uni}, t}], \\
    \bm{o}^{\rm H}_t[{\rm grasping\ states}] &:= \bm{\theta}^{\text{gripper}}_{t}, \\
    \bm{a}^{\rm H}_t[{\rm body\ pose}] &:=[\bm{x}^{\text{head, t}}_{\text{uni}, t}, \bm{R}^{\text{head, t}}_{\text{uni}, t}], \\
    \bm{a}^{\rm H}_t[{\rm EEF\ pose}] &:=[\bm{x}^{\text{r-wrist, t}}_{\text{uni}, t}, \bm{R}^{\text{r-wrist, t}}_{\text{uni}, t}, \\ 
    & \qquad \bm{x}^{\text{l-wrist, t}}_{\text{uni}, t}, \bm{R}^{\text{l-wrist, t}}_{\text{uni}, t}], \\
    \bm{a}^{\rm H}_t[{\rm grasping\ actions}] &:= \bm{\theta}^{\text{gripper, t}}_{t} \label{eq:humandata}
\end{aligned}
\end{equation}

In this way, we ensure that the human and robot data are unified in terms of both format and spatial interpretation, and can be used to train our proposed Modularized Cross-Embodiment Transformer introduced in Section \ref{sec:mxt}.

\subsection{Modularized Cross-embodiment Transformer}
\label{sec:mxt}


Given our unified multi-embodiment data collection pipeline, we aim to train a cross-embodiment policy where the overall structure and the majority of parameters are transferrable, while accounting for modality-specific distributions unique to each embodiment. To this end, we propose a modularized design called \textit{\textbf{M}odularized \textbf{Cross}-embodiment \textbf{T}ransformer} (MXT). As illustrated in Figure \ref{fig:arch}, MXT consists mainly of three groups of modules: tokenizers, Transformer trunk, and detokenizers. The tokenizers act as encoders and map embodiment-specific observation modalities to tokens in the latent space, and the detokenizers translate the output tokens from the trunk to action modalities in the action space of each embodiment. The tokenizers and detokenizers are specific to one embodiment and are reinitialized for each new embodiment, while the trunk is shared across all embodiments and reused for transferring the policy among embodiments. 

\parab{Tokenizers.} The tokenizers $T$ transform raw observations into tokens for the Transformer trunk. Similar to the design in \cite{wang2024scaling}, we use a cross-attention layer to format observational features into a fixed number of tokens for each modality. For image inputs, the features are obtained from a pretrained ResNet encoder that can be finetuned during training; for proprioceptive or state-like inputs, the features are computed by passing the raw input through a trainable MLP network.

\parab{Detokenizers.} The detokenizers $D$ serve as action decoder heads, mapping the output tokens from the trunk to the action modalities in each embodiment’s action space. To reduce compounding errors and lower inference frequency, we adopt action chunking~\cite{zhao2023learning}, where the detokenizers predict a sequence of $h$ actions at each inference step. Within each detokenizer, a cross-attention layer is used to transform the latent action tokens—produced by the trunk at fixed positions—into a sequence of $h$ actions matching the dimensions of the corresponding action modality. 

\parab{Trunk.} The trunk is an encoder-decoder Transformer, where the input sequence length and the output sequence length are both fixed, as the number of tokens for each observation or action modality is fixed by design. By sharing the trunk weights across human and robot embodiments, the model learns to process and generate tokens of different modalities and capture common decision-making patterns across embodiments.

\parab{Modality Decomposition in Tokenizers / Detokenizers.}
Due to the aligned data format and the unified observation and action spaces across embodiments, we are able to separately transform each semantically distinct component of the observational input and the action output, which we refer to as \textit{modality}, and specify the compositional structure at the interface of the Transformer trunk and the tokenizers / detokenizers. This design preserves modality-specific distributions unique to each embodiment and enables the model to explicitly account for distributional gaps across embodiments, which is core to the effectiveness of our method.

Concretely, for tokenization in the embodiment $e$, we encode the input observation $\bm{o}_t$ with multiple tokenizers $\{T_{e,m_i}\}$ at the finer granularity of modalities denoted by $\bm{o}_t[m_i]$. For instance, instead of aggregating all image inputs before passing through the vision tokenizer, we use separate tokenizers for the main camera and the wrist camera views
. All the encoded modalities are concatenated to compose the input tokens to the Transformer trunk.

Similarly, for detokenization, we specify the subset of the Transformer output tokens corresponding to each action modality, e.g. body pose, end effector pose, and gripper actions, and decode the selected tokens to yield each modality with separate detokenizers $\{D_{e, m_i}\}$. For convenience, we use the set of observation and action modalities as defined by the data collection formats in (\ref{eq:robotdata}) and (\ref{eq:humandata}).

By explicitly decomposing the input and output modalities and encoding them separately, we are leveraging the innate structure of observations and actions and imposing such a structure on the token sequences processed by the Transformer. Consequently, the knowledge of how to process different modalities learned during training can be shared across embodiments, fostering efficient transfer of the policy.

Although we employ a consistent data format and aligned input/output representations across embodiments, some modalities are not present or available for all embodiments. For example, the human operator is not equipped with a wrist camera, while the LocoMan robot has a wrist camera in some tasks to improve manipulation accuracy. In this case, we use masks defined during data collection to signify redundant dimensions in the observations as well as in the action labels. We refer the reader to Appendix Section \ref{sec:MXT para} for more implementation details.

In general, the highly modularized design of our learning framework offers great flexibility in handling all types of manipulation tasks across different embodiments, and effectively enhances the learning performance by capturing the common patterns in manipulation problems.



\begin{table*}[htp]
\centering
\caption{\oursystem embodiments (R$=$Right, L$=$Left).}
\vspace{-0.1cm}
\label{tab:embodiments}
\resizebox{1.36\columnwidth}{!}{%
\begin{tabular}{l|c c c c c c c c c c}
\toprule
\multirow{2}{*}{Embodiments} & Head & Wrist & Body & R-EEF & L-EEF & Body & R-EEF & L-EEF & R-Grasp & L-Grasp \\
 & Images & Image & Priop. & Priop. & Priop. & Pose & Pose & Pose & Action & Action \\
\midrule
Human-Unimanual (R) & \checkmark & $\times$ & \checkmark & \checkmark & $\times$ & \checkmark & \checkmark & $\times$ & \checkmark & $\times$ \\
Human-Unimanual (L) & \checkmark & $\times$ & \checkmark & $\times$ & \checkmark & \checkmark & $\times$ & \checkmark & $\times$ & \checkmark \\
Human-Bimanual & \checkmark & $\times$ & \checkmark & \checkmark & \checkmark & \checkmark & \checkmark & \checkmark & \checkmark & \checkmark \\
LocoMan-Unimanual (R) & \checkmark & \checkmark & \checkmark & \checkmark & \checkmark & \checkmark & \checkmark & $\times$ & \checkmark & $\times$ \\
LocoMan-Unimanual (L) & \checkmark & \checkmark & \checkmark & \checkmark & \checkmark & \checkmark & $\times$ & \checkmark & $\times$ & \checkmark \\
LocoMan-Bimanual & \checkmark & $\times$ & \checkmark & \checkmark & \checkmark & $\times$ & \checkmark & \checkmark & \checkmark & \checkmark \\
\bottomrule
\end{tabular}%
}
\end{table*}

\subsection{Training Paradigm}

The details of our MXT training pipeline are outlined in Algorithm~\ref{algo:train}. For a given task, we first pretrain the model using the human dataset, followed by finetuning with the corresponding LocoMan dataset. During finetuning, only the Transformer trunk weights are initialized from the pretrained checkpoint. For tasks that share similar semantics but differ in manipulation modes (representing distinct embodiments in Table~\ref{tab:embodiments}), we jointly pretrain the model on human datasets across these tasks with different manipulation modes and then finetune it on each task using the corresponding LocoMan robot dataset.

\parab{Learning Objective.} We use the behavioral cloning objective for both pretraining and finetuning. In general, given a dataset $\mathcal{D}_e$ on an embodiment $e$ and aligned action modalities $m_1,...,m_k$, the total loss to optimize when training on $e$ is:
 \begin{align}
     \mathcal{L}_{e}(\theta)= \sum_{i=1}^k \mathcal{L}_{e, m_i}(\theta), \label{eq:training}
\end{align}
 where $\mathcal{L}_{e,m_i}$ is the $\ell_1$ loss of the action modality $m_i$ with respect to the dataset of embodiment $e$.
In practice, we optimize the following batched loss for each training batch $B_e=\{(\bm{o}_j, {A_j})\}_{j=1}^n$ as a proxy of $\mathcal{L}_{e, m_i}(\theta)$:
\begin{align}
    \mathcal{L}_{e, m_i}(B_e)=\frac{1}{n}\sum_{j=1}^n\left[\frac{1}{h}\sum_{l=1}^h\ell_1\left(\bm{a}_{j,l}\left[m_i\right],\hat{\bm{a}}_{j,l}\left[m_i\right]\right)\right], \label{eq:batch_loss}
\end{align}
where $\bm{a}_{j,l}\left[m_i\right]=(A_j)_{l}\left[m_i\right]$ is the $l$-th step action of modality $m_i$ in the action label sequence sample $A_j=\left\{\bm{a}_{j,l}\right\}_{l=1}^h$; $\hat{\bm{a}}_{j,l}\left[m_i\right]=\left[\pi_\theta(\bm{o}_j)\right]_{l}\left[m_i\right]$ is the predicted action of modality $m_i$ at step $l$, and $h$ is the chunk size or the action horizon.

\begin{algorithm}
\caption{Pretraining MXT on human data and finetuning on LocoMan data}\label{algo:train}
    \begin{algorithmic}
        \Require Human dataset $\mathcal{D}_{\rm human}$, LocoMan dataset $\mathcal{D}_{\rm LocoMan}$
        \Ensure Policy $\pi$ for versatile LocoMan manipulation
        \State Initialize the MXT policy network $\pi_\theta$ with parameters $\theta$.
        \State Set pretraining learning rate $\eta_{\rm pretrain}$
        \For{step $=1,2,...$} \Comment{Pretraining Stage}
            \State Sample a batch $B$ from $\mathcal{D}_{\rm human}$
            \State Compute $\mathcal{L}_{\text{human}}(B)=\sum_{i}\mathcal{L}_{\text{human}, m_i}(B)$ with Eq.\ref{eq:batch_loss}
            \State Optimize the policy weights $\theta$ with backpropagation
        \EndFor
        \State Reinitialize the tokenizers and detokenizers of $\pi$. Preserve the trunk weights $\theta_{\rm trunk}$ learned from pretraining.
        \State Set finetuning learning rate $\eta_{\rm finetune}$
        \For{step $=1,2,...$} \Comment{Finetuning Stage}
            \State Sample a batch $B$ from $\mathcal{D}_{\rm LocoMan}$
            \State Compute $\mathcal{L}_{\text{LocoMan}}(B)=\sum_{i}\mathcal{L}_{\text{LocoMan}, m_i}(B)$ with Eq.\ref{eq:batch_loss}
            \State Optimize the policy weights $\theta$ with backpropagation
            
        \EndFor
        \State \Return $\pi$
    \end{algorithmic}
\end{algorithm}

%% file: arxiv_contents/experiments.tex
\begin{figure*}
    \centering
    \includegraphics[width=1.00\linewidth]{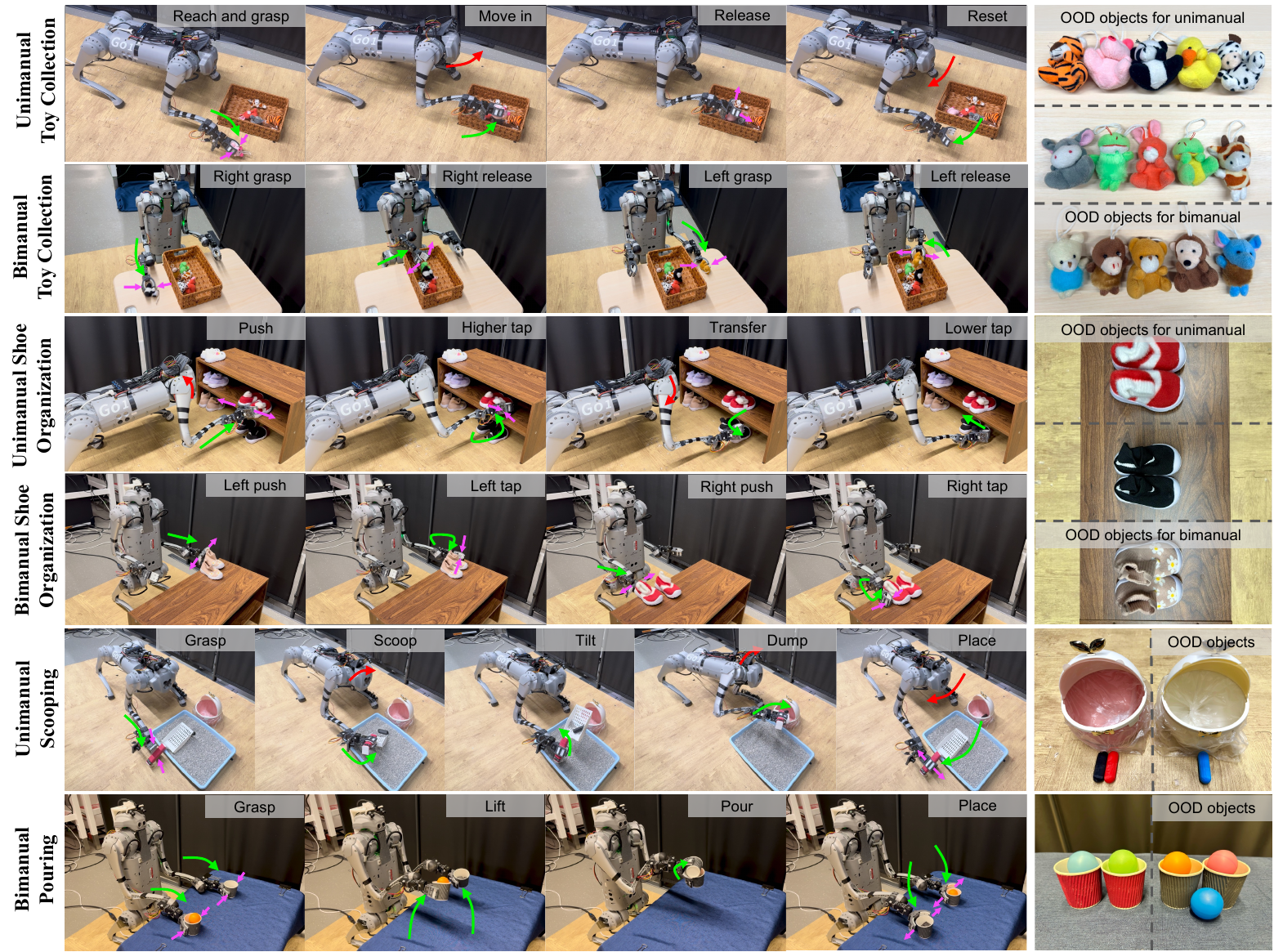}
    \caption{Rollouts of the MXT policy and the objects used across manipulation tasks in our experiments. \textcolor{eef_motion}{Green arrows} indicate end-effector motions, \textcolor{body_motion}{red arrows} denote torso movements, and \textcolor{gripper_action}{pink arrows} represent gripper actions. Both \textbf{unimanual} and \textbf{bimanual toy collection} tasks assess the robot’s ability to grasp objects of varying shapes, colors, and positions. The unimanual variant emphasizes coordination between the torso and end-effector, while the bimanual variant highlights synchronized control of two loco-manipulators. \textbf{Unimanual} and \textbf{bimanual shoe rack organization} tasks evaluate non-prehensile manipulation skills such as pushing and tapping. The unimanual variant additionally requires torso articulation to reach shoes placed at different heights. \textbf{Scooping} is a complex task involving tool use, deformable object manipulation, and wide-range torso motion. \textbf{Pouring} is a long-horizon task that demands precise coordination of both loco-manipulators.}
    \label{fig:exp_settings}
\end{figure*}

\section{Experiments}
\label{sec:experiments}


In this section, we aim to answer the following research questions: (1) Does the \oursystem system enable versatile quadrupedal manipulation capabilities? (2) How does MXT compare to state-of-the-art imitation learning architectures? (3) How does human data collected by \oursystem contribute to imitation learning performance? (4) Do the design choices in MXT facilitate positive transfer from Human to LocoMan?


\subsection{Experimental Setup}

\subsubsection{Tasks}
We evaluate MXT on six household tasks of varying difficulty, across unimanual and bimanual manipulation modes of the LocoMan robot, with data collected by the \oursystem system:

\begin{itemize}
    \item \parai{Unimanual Toy Collection (TC-Uni).} In this task, the robot must pick up a toy randomly positioned within a rectangular area and place it into a designated basket on the ground. Completing this task requires the robot to coordinate its whole-body motions to efficiently and accurately reach various locations on the ground and above the basket. As shown in Figure~\ref{fig:exp_settings}, we use 10 objects for robot finetuning and all objects for human pretraining and real-robot evaluation. The substeps of this task include: grasp the toy, and release the toy.
    \item \parai{Bimanual Toy Collection (TC-Bi).} Similar to \parai{Unimanual Toy Collection}, this task requires the robot to pick up a toy randomly placed within two rectangular areas on either side of a basket. We use 10 objects for robot finetuning, while all objects are included in human pretraining and real-robot evaluation. The substeps of this task include: grasp the toy, and release the toy.
    \item \parai{Unimanual Shoe Rack Organization (SO-Uni).}  
    This longer-horizon task involves organizing two shoes placed on different levels of a shoe rack. The robot must coordinate whole-body motions to reach various rack levels and utilize both prehensile and non-prehensile manipulation skills. As shown in Figure~\ref{fig:exp_settings}, this task involves three pairs of shoes, with one pair being out-of-distribution (OOD). The substeps of this task include: push the shoe on the higher rack, tap the shoe on the higher rack, transfer the gripper to the lower level, and tap the shoe on the lower rack.
    \item \parai{Bimanual Shoe Rack Organization (SO-Bi).}  
    One pair of shoes is randomly placed at the edge of the third level of the shoe rack. The robot must push one shoe inward and align it with the other. The substeps of this task include: push the shoe, and tap the shoe.
    \item \parai{Unimanual Scooping (Scoop-Uni).} The robot performs unimanual manipulation using a litter shovel to scoop a 3D-printed cat litter from varying locations and poses within a litter box, and then dump it into a trash bin. This long-horizon task involves both tool use and deformable object manipulation. The task is decomposed into the following substeps: grasp the shovel, scoop the litter, tilt the shovel, dump the litter, and place the shovel back.
    \item \parai{Bimanual Pouring (Pour-Bi).} The robot performs bimanual manipulation to pour a Ping Pong ball from one cup to another. This longer-horizon task requires the robot to accurately reach both cups, which are randomly placed within a rectangular area on a table, lift one cup, pour the ball into the other, and then place both cups back on the table. This task evaluates the coordination and precision of the robot's bimanual manipulation. The substeps of this task include: pick up both cups, pour the ball, and place both cups.
\end{itemize}

\subsubsection{\oursystem Embodiments}
As shown in Table~\ref{tab:embodiments}, the unimanual and bimanual modes of \oursystem represent distinct embodiments, each differing in morphology,
observations, and action spaces. In practice, we install and utilize wrist cameras on the LocoMan robot for the three unimanual manipulation tasks.

\subsubsection{Data collection}
For each task, we collect various numbers of human and robot trajectories with the \oursystem system. The details of the collected data are demonstrated in Table \ref{Tab:data}. About $10\%$ data of each task is used for validation.

\subsubsection{Training details.} For Toy Collection and Shoe Rack Organzation, we pretrain a model that utilizes the human data of both the unimanual and bimanual versions of the task, then we finetune the model on each unimanual or bimanual task with the corresponding robot data. For each task, we choose a set of training hyperparameters (e.g. batch size, chunk size) that are kept the same for all methods. (See Appendix Section \ref{sec:global para}.) We also list the model hyperparameters we use for our method and the baselines in the Appendix Section \ref{sec:MXT para} and \ref{sec:baseline para}.

\input{arxiv_tables/main_table}

\begin{table}[htp]
\caption{Records of data collection for different tasks.}
\label{Tab:data}
\centering
\resizebox{0.92\columnwidth}{!}{
\begin{tabular}{@{}lccccc@{}}
\toprule
Task         & \# human traj.    & human time (min)  & \# robot traj.  & robot time (min) \\ \midrule
TC-Uni       & 300               & 25                & 150             & 15              \\
TC-Bi        & 315               & 22                & 70              & 7              \\
SO-Uni       & 240               & 34                & 90              & 23              \\
SO-Bi        & 200               & 20                & 92              & 12              \\
Scoop-Uni & 340               & 96                & 66              & 22              \\
Pour-Bi   & 210               & 35                & 64              & 22              \\ \bottomrule
\end{tabular}
}
\end{table}

\subsubsection{Baselines} 
We compare \oursystem to the following SOTA imitation learning baselines:
\begin{itemize}
    \item \textit{Humanoid Imitation Transformer (HIT):} HIT~\cite{fu2024humanplus} is an imitation learning framework originally developed for humanoid skill learning, with the capability to generalize to other robot embodiments. It builds on ACT~\cite{zhao2023learning} and employs a decoder-only architecture that simultaneously predicts future action sequences and future image features. To prevent the vision-based policy from ignoring visual inputs and overfitting to proprioceptive states, HIT introduces an L2 loss on image features in addition to the standard behavioral cloning objective. Empirically, HIT consistently outperforms ACT across our evaluated tasks. While it is not designed to handle data from diverse domains or embodiments, we position HIT as a strong reference implementation for efficient imitation learning from in-domain robot demonstrations.
    \item \textit{Heterogeneous Pretrained Transformer (HPT):} HPT~\cite{wang2024scaling} is a framework that pretrains a policy on heterogeneous datasets comprising simulation data, real robot trajectories, and human videos. It consists of stems, a trunk, and a head, where the stems and head serve roles analogous to our tokenizers and detokenizers. The trunk is designed to learn the complex mapping between inputs and outputs within a unified latent space through large-scale pretraining.
    The implementation of HPT differs from our framework in several aspects. First, we align data at the modality level and design the modular architecture in MXT to preserve modality-specific information across embodiments. In contrast, HPT uses a single tokenizer for visual inputs and another for proprioceptive inputs, along with a single detokenizer for all action dimensions. Additionally, HPT freezes its ResNet image encoder, whereas we finetune the ResNet encoder together with the rest of the network in an end-to-end manner for better adaptation to a specific embodiment.
\end{itemize}
More implementation details of these baselines can be found in Appendix Section \ref{sec:baseline para}. For the HPT baseline, we train with several different settings: training with only LocoMan data, pretraining with our human data and finetuning on LocoMan data, and directly finetuning the released HPT checkpoints with LocoMan data. For the HIT baseline, we only train on LocoMan data, as it is unable to incorporate human data.

\subsubsection{Evaluation Metrics}
We present the evaluation results using three metrics: i) success rate (SR), ii) task score (TS), and iii) validation loss.
To calculate the success rate and task score, we perform a fixed number of real world rollouts using the evaluated method for one task. The policy is rolled out for 24 times with in-distribution (ID) objects and 12 times with out-of-distribution (OOD) objects. 

For each task, we define a set of critical substeps necessary to fully complete the task. When calculating the task score, successfully completing each intermediate substep earns one point, and reaching the final goal—i.e., completing the entire task—earns an additional point. The final task score is the sum of points across all rollouts for that task. The success rate of a method on a given task, under either the ID or OOD setting, is computed as the ratio of successful rollouts (i.e., rollouts where all substeps are completed) to the total number of rollouts performed.


In addition, we report the best validation loss as another metric for training performance. For all the included methods, we align how the loss is computed so that these losses can be meaningfully compared. Note that the validation loss is not a faithful indicator of the policy performance, but it does reflect how well the model is optimized, especially when there is a significant difference in the validation loss of different policies in the same setting. We mainly use this metric to analyze the training process of different architectures (MXT, HIT and HPT) and to provide a separate dimension to our evaluation.

\begin{figure*}
    \centering
    \includegraphics[width=0.92\linewidth]{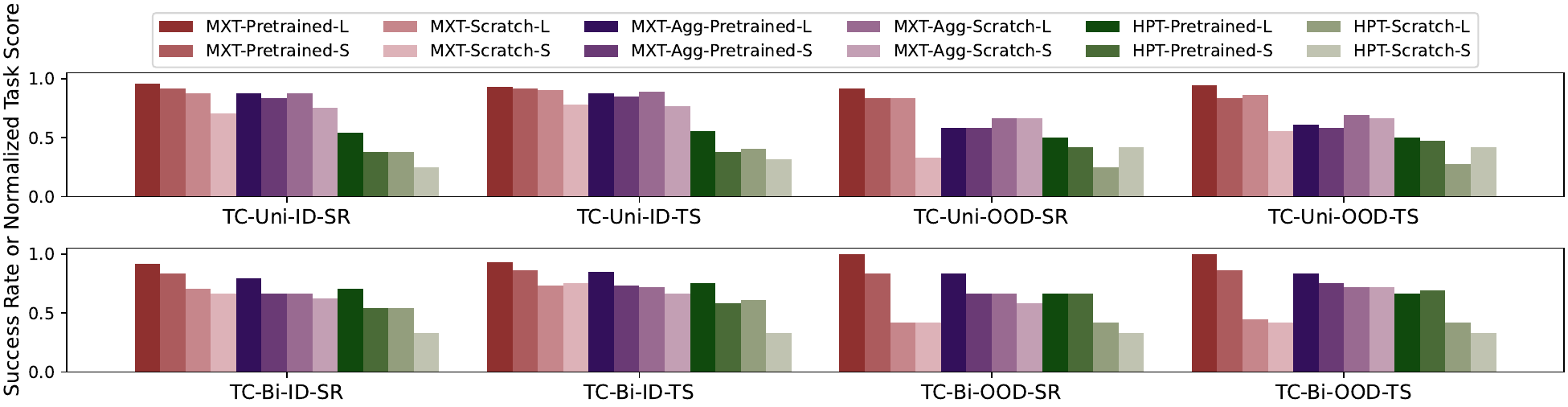}
    \caption{Ablation study on unimanual and bimanual toy collection. We compare MXT, its ablation MXT-Agg, and baseline HPT on SR and TS. Here, ``L'' denotes the larger training set (40 trajectories for TC-Uni, 60 trajectories for TC-Bi), while ``S'' denotes the smaller training set (20 trajectories for TC-Uni, 30 trajectories for TC-Bi).}
    \label{fig:ablation}
\end{figure*}

\subsection{Results and Analysis}

\parai{(1) Does the \oursystem system enable versatile quadrupedal manipulation capabilities?}

\parab{Data collection.} As shown in Table~\ref{Tab:data}, \oursystem teleoperation enables the collection of a substantial amount of robot data (over 50 trajectories) within 30 minutes across all tasks. Using the \oursystem human data collection system, over 200 trajectories can be gathered within the same time frame. Even for the most challenging task, a human can collect over 300 trajectories within one and a half hours. Notably, the robot’s manipulation speed is comparable to, and in many tasks approaches, that of a human. These results highlight the data collection efficiency of our system.

\parab{Task versatility.} As depicted in Figure~\ref{fig:exp_settings}, \oursystem's policy can perform tasks across a wide range of scenarios, including unimanual and bimanual manipulation, prehensile and non-prehensile manipulation, deformable object manipulation, and tool use, while also generalizing to OOD objects and conditions. 

\parab{Task performance.} We summarize the success rates and task scores of our method and HIT across all tasks in Table~\ref{table:main}. \oursystem's MXT presents strong performance using a relatively small robot dataset, achieving success rates above $79\%$ across all tasks. The baseline method also attains decent performance on most tasks. These results highlight the high quality of our collected data and demonstrate the effectiveness of \oursystem's data collection and training pipeline.

\begin{figure}
    \centering\includegraphics[width=\columnwidth]{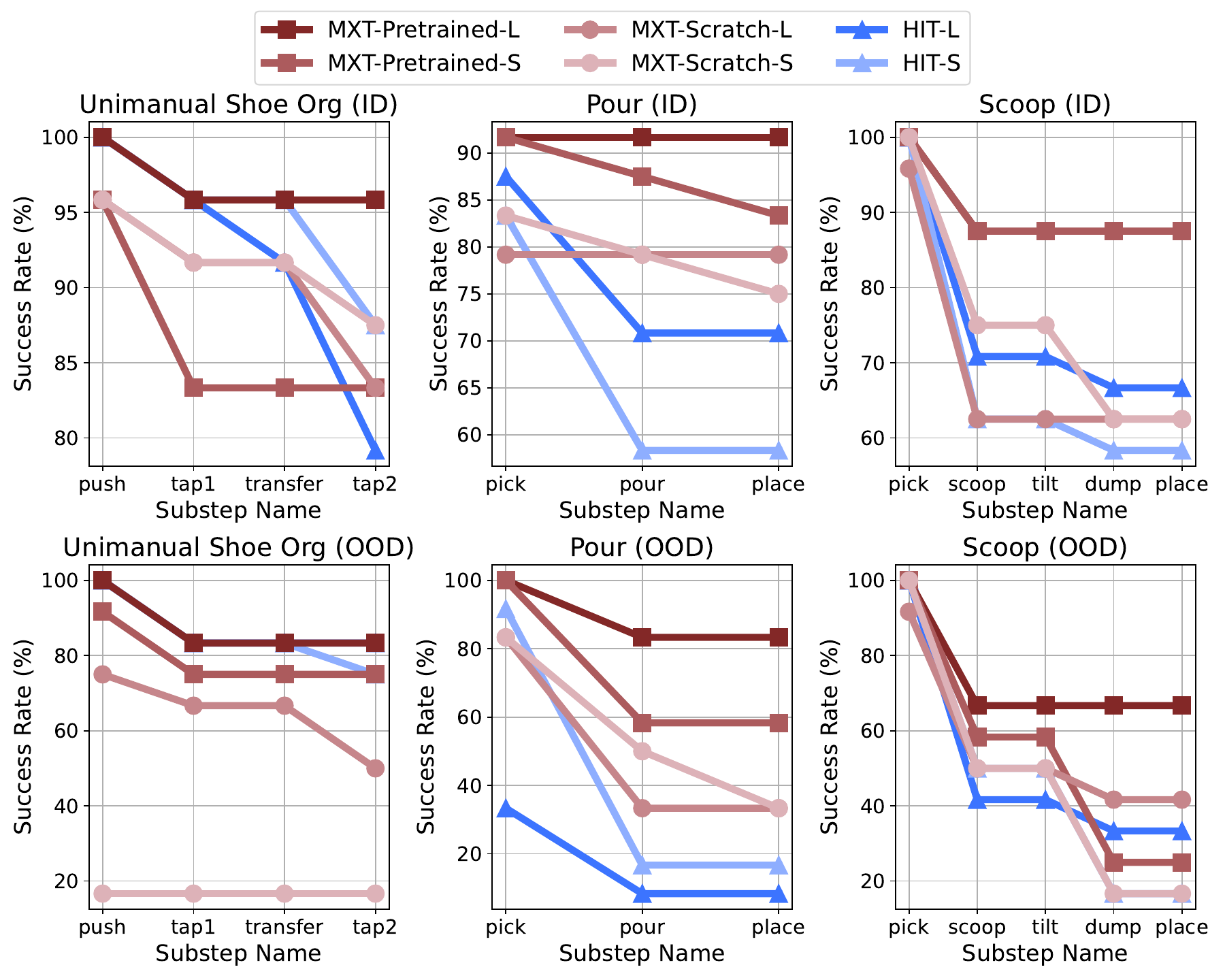}
    \caption{Substep success rate. The success rate for some substep is calcuated as the percentage of trials where the robot successfully completed the substep. For each task, we calculate this with 24 ID rollouts and 12 OOD rollouts. \textbf{MXT-Pretrained}: MXT pretrained on human dataset (including unimanual and bimanual if applicable), then finetuned on the LocoMan data. \textbf{MXT-Scratch}: MXT trained only on the LocoMan data. ``L'' denotes the larger training set (80 trajectories for SO-Uni, 60 trajectories for Pour and Scoop), while ``S'' denotes the smaller training set (40 trajectories for SO-Uni, 30 trajectories for Pour and Scoop).}
    \label{fig:substep_sr}
\vspace{-4mm}
\end{figure}

\parai{(2) How does MXT compare to state-of-the-art imitation learning architectures?}

\begin{figure}
    \centering\includegraphics[width=\columnwidth]{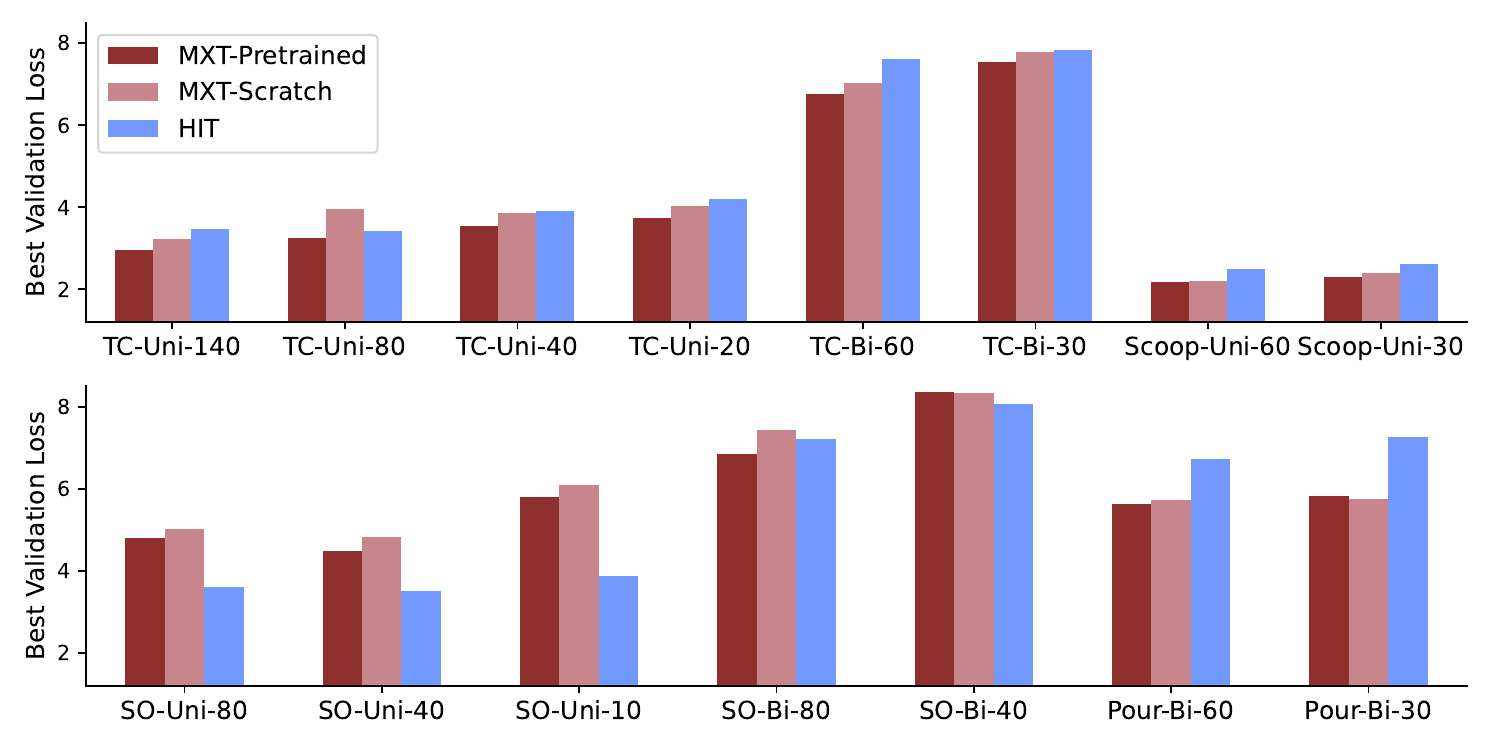}
    \caption{Best validation loss of our method and HIT on all our tasks. \textbf{MXT-Pretrained}: MXT pretrained on human dataset (including unimanual and bimanual if applicable), then finetuned on the LocoMan data. \textbf{MXT-Scratch}: MXT trained only on the LocoMan data. The number suffix denotes the number of demonstrations in the LocoMan training set.}
    \label{fig:val_loss}
\end{figure}

\parab{Compared to HIT.} 
As shown in Table \ref{table:main}, in most evaluated tasks, spanning both unimanual and bimanual modes and across both ID and OOD inference scenarios, MXT without pretraining achieves comparable or superior performance relative to HIT. Moreover, pretrained MXT consistently outperforms the HIT baseline in terms of both SR and TS. 
Specifically, pretrained MXT achieves an average SR improvement of $41.9\%$ overall and $79.7\%$ under OOD settings, as well as an average TS improvement of $28.9\%$ overall and $51.4\%$ under OOD settings.
From Figure~\ref{fig:val_loss}, we find that MXT demonstrates lower validation loss compared to HIT on most tasks, indicating superior training convergence. The performance improvement is particularly evident in tasks with larger datasets, suggesting that MXT scales more effectively with increasing data availability. Notably, HIT achieves a significantly lower validation loss compared to the MXT variants in unimanual shoe organization, while attaining comparable performance in SR and TS metrics under both ID and OOD settings relative to the best MXT model. As shown in the substep success analysis in Figure~\ref{fig:substep_sr}, the primary failures of the lower-performing MXT models occur during the first two substeps, push'' and tap1.'' One potential reason for this is that the unimanual shoe organization task exhibits relatively less variation in object locations and types compared to other tasks, which may favor HIT despite its lack of modular designs and pretraining.

\begin{figure}
    \centering\includegraphics[width=0.99\linewidth]{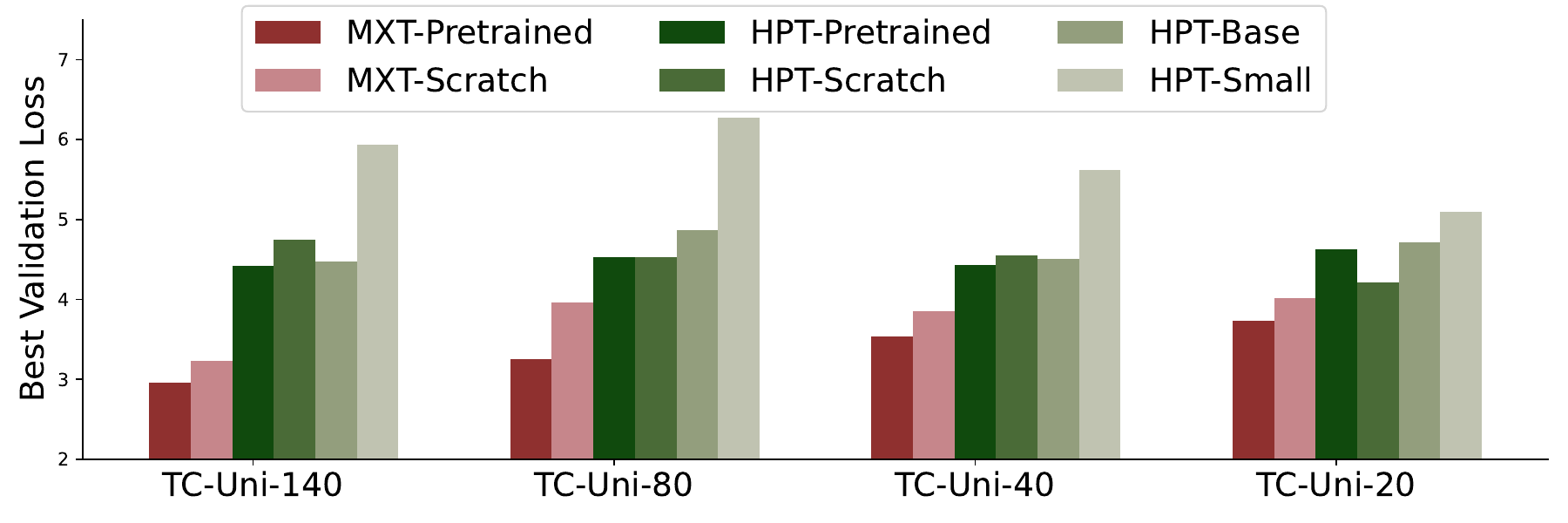}
    \caption{Best validation loss of our method and HPT on the unimanual Toy Collection task. \textbf{MXT-Pretrained}: MXT pretrained on human dataset (including unimanual and bimanual if applicable), then finetuned on the LocoMan data. \textbf{MXT-Scratch}: MXT trained only on the LocoMan data. \textbf{HPT-Pretrained}: HPT trunk pretrained on our human data, then finetuned on the LocoMan data. \textbf{HPT-Scratch}: HPT network trained only on the LocoMan data. \textbf{HPT-Base}: Finetune with our LocoMan data with HPT trunk initialized with released \texttt{HPT-Base} weights. \textbf{HPT-Small}: Finetune with our LocoMan data with HPT trunk initialized with released \texttt{HPT-Small} weights.}
    \label{fig:val_loss_hpt}
\vspace{-4mm}
\end{figure}

\parab{Compared to HPT.}
As shown in Figure~\ref{fig:ablation}, we present SR and TS results based on 36 trials, comprising 24 OOD and 12 ID trials. HPT consistently underperforms compared to MXT, both when finetuned and when trained from scratch, across all data sizes on the toy collection tasks. We attribute this performance gap to a combination of HPT’s lack of modular architecture and its use of frozen image encoders. 
Validation loss results, shown in Figure~\ref{fig:val_loss_hpt}, reveal a similar trend in the unimanual toy collection task across a broader range of dataset sizes.
Notably, HPT-Small and HPT-Base fail to achieve lower validation loss compared to HPT-Scratch, even pretrained on the large public dataset. Their policy rollout performance are evidently worse than HPT-Scratch and HPT-Pretrained so we did not scale up the real-robot evaluation. This indicates the challenge posed by the large embodiment gap and underscores the effectiveness of the human data collected by our system. 
Furthermore, we observe severe overfitting in HPT experiments when training on our datasets, a phenomenon not observed in MXT. This further suggests that the modular design of the MXT architecture facilitates better generalization.


\parai{(3) How does human data collected by \oursystem contribute to imitation learning performance?} 

\parab{Efficiency, robustness, and generalizability.} 
Summarizing from Table~\ref{table:main}, pretraining with human data improves the MXT policy SR by $38.6\%$ overall and $82.7\%$ under OOD settings, and boosts TS by $24.1\%$ overall and $58.9\%$ under OOD settings. This enables consistently stronger performance with only half the amount of robot data, demonstrating both the efficiency and robustness of our system.
We hypothesize that MXT benefits from learning useful complementarities—i.e., positive transfer effects—between human demonstrations and LocoMan robot data. Specifically, comparing MXT-Pretrained to MXT-Scratch in Table~\ref{table:main}, we observe that pretraining improves performance on TC-Uni, TC-Bi, and Scooping tasks under ID settings, where objects exhibit diverse \textbf{locations}. MXT-Pretrained tends to produce smoother and more robust motions, enabling more accurate localization of target objects. For instance, as shown in Figure~\ref{fig:substep_sr}, MXT-Pretrained achieves substantially better scooping performance—which requires precise localization—compared to all other methods.
Moreover, Table~\ref{table:main} reveals large performance gaps on OOD objects in tasks such as TC-Bi, SO-Uni, and Pouring, where OOD objects differ significantly from ID objects in \textbf{shape}, \textbf{texture}, and \textbf{color}. These results suggest that MXT, by leveraging human demonstrations during the pretraining stage, is able to generalize effectively to novel scenarios unseen during robot training.


\parab{Long-horizon performance.} For a more detailed analysis on long-horizon tasks that require multiple manipulation steps, we present in Figure~\ref{fig:substep_sr} how the success rate decays with each substep in tasks including SO-Uni, Pour-Bi and Scoop-Uni. MXT-Pretrained is shown to maintain a decent success rate as the long-horizon task progresses, while MXT-Scratch and HIT tend to fail more after the first substep, especially in Pouring and Scooping tasks. We note that the second substep in these tasks commonly involves moving and localizing an object with precision, and pretraining with human data appears to help with completing such challenging substeps. This suggests that human data incorporated during pretraining can promote manipulation accuracy, which is key to completing a sequential long-horizon task.

\parai{(4) Do the design choices in MXT facilitate positive transfer from Human to LocoMan?}
We have shown in Figure~\ref{fig:ablation} that MXT outperforms HPT under both finetuning and training-from-scratch settings, benefiting from its modularized design and unfrozen image encoders in its architecture. Here, we observe the delta performance brought by human data across different methods.
While both MXT and HPT benefit from pretraining on human data, MXT exhibits more effective Human-to-LocoMan transfer, achieving larger gains,
especially in low-data and OOD settings. Additionally, as depicted in Figure 8, MXT-Pretrained consistently achieves lower validation loss than MXT-Scratch, whereas the gap between HPT-Pretrained and HPT-Scratch is less consistent and does not always show the same trend. These results highlight the ability of MXT to consume human data, despite the large embodiment gap with LocoMan.

To further investigate the impact of modularity, we introduce an ablation variant of MXT, referred to as MXT-Agg, in which we \textit{aggregate} the input modalities: a single visual tokenizer is used to encode all visual observations, a single proprioceptive tokenizer for all proprioceptive inputs, and a single detokenizer for all action dimensions—mirroring HPT’s design. MXT-Agg incorporates HPT’s key features, including cross-attention tokenization and trunk weight sharing, while still finetuning the vision encoders and remaining architecturally comparable to MXT. Across evaluations, MXT consistently benefits from pretraining and outperforms MXT-Agg when both are finetuned, highlighting the advantages of modularized tokenization for effectively leveraging human data.
Notably, MXT-Agg exhibits weaker transfer performance with respect to MXT and HPT, as evidenced by little to no improvement when finetuning the pretrained model compared with training from scratch. This is likely due to increased representation power in the tokenizer, which permits more overfitting in the transformer trunk and could negatively impact the trunk transferability. However, with the incorporation of our modular design, MXT is trained with additional regularization and exhibits improved transferability. The modular design effectively aids in the trade-off between more network representation power and better transferability in our framework, and allows attaining both qualities.

%% file: arxiv_tables/main_table.tex
\colorlet{HIT-S-table}{HIT-S!50}
\colorlet{HIT-L-table}{HIT-L!50}
\colorlet{MXT-Scratch-S-table}{MXT-Scratch-S!40}
\colorlet{MXT-Scratch-L-table}{MXT-Scratch-L!40}
\colorlet{MXT-Pretrained-S-table}{MXT-Pretrained-S!40}
\colorlet{MXT-Pretrained-L-table}{MXT-Pretrained-L!40}

\begin{table*}[h]
\centering
\caption{Result Summary. We report success rate (SR) $\uparrow $ in \% and task score (TS) $\uparrow$ for each task. We highlight the best performance in \textbf{bold} and the second best in \underline{underline}. ID results are based on 24 trials, and OOD results on 12 trials.}
\vspace{-4mm}
\begin{center}
\begin{adjustbox}{width=1.0\linewidth}
\begin{tabular}{lcr|cccc|cccc|cccc|cccc|cccc|cccc}
\toprule 
&& & \multicolumn{8}{c}{\textbf{Toy Collection}} & \multicolumn{8}{c}{\textbf{Shoe Rack Organization}} & \multicolumn{4}{c}{\textbf{Scooping}} & \multicolumn{4}{c}{\textbf{Pouring}} \\   
  \cmidrule(lr){4-11} \cmidrule(lr){12-19}\cmidrule(lr){20-23}\cmidrule(lr){24-27} 
&&& \multicolumn{4}{c|}{Unimanual} &\multicolumn{4}{c|}{Bimanual} & \multicolumn{4}{c|}{Unimanual} &\multicolumn{4}{c|}{Bimanual} & \multicolumn{4}{c|}{Unimanual} & \multicolumn{4}{c}{Bimanual} \\
&& & \multicolumn{2}{c}{ID}  & \multicolumn{2}{c|}{OOD} &  \multicolumn{2}{c}{ID}  & \multicolumn{2}{c|}{OOD}  & \multicolumn{2}{c}{ID}  & \multicolumn{2}{c|}{OOD}  & \multicolumn{2}{c}{ID}  & \multicolumn{2}{c|}{OOD}  & \multicolumn{2}{c}{ID}  & \multicolumn{2}{c|}{OOD} & \multicolumn{2}{c}{ID}  & \multicolumn{2}{c}{OOD} \\
\textbf{Method}&Pretrained&Data & SR & TS & SR & TS& SR & TS& SR & TS& SR & TS& SR & TS& SR & TS& SR & TS& SR & TS& SR & TS & SR & TS& SR & TS
\\\toprule

\textcolor{HIT-S}{\textbf{HIT}}  & \textcolor{HIT-S}{\textbf{-}} &\textcolor{HIT-S}{\textbf{smaller}}    
& \cellcolor{HIT-S-table}54.2 & \cellcolor{HIT-S-table}42 
& \cellcolor{HIT-S-table}41.6 & \cellcolor{HIT-S-table}15 
& \cellcolor{HIT-S-table}45.8 & \cellcolor{HIT-S-table}37
& \cellcolor{HIT-S-table}41.6 & \cellcolor{HIT-S-table}16 
& \cellcolor{HIT-S-table}\underline{87.5} & \cellcolor{HIT-S-table}\underline{112}
& \cellcolor{HIT-S-table}75.0   & \cellcolor{HIT-S-table}50 
& \cellcolor{HIT-S-table}66.7 & \cellcolor{HIT-S-table}52 
& \cellcolor{HIT-S-table}25.0 & \cellcolor{HIT-S-table}14 
& \cellcolor{HIT-S-table}58.3 & \cellcolor{HIT-S-table}96
& \cellcolor{HIT-S-table}16.7 & \cellcolor{HIT-S-table}30
& \cellcolor{HIT-S-table}58.3 & \cellcolor{HIT-S-table}62
& \cellcolor{HIT-S-table}16.7 & \cellcolor{HIT-S-table}17 
\\ 
\textcolor{HIT-L}{\textbf{HIT}}  & \textcolor{HIT-L}{\textbf{-}} &\textcolor{HIT-L}{\textbf{larger}}       
& \cellcolor{HIT-L-table}79.2 & \cellcolor{HIT-L-table}57
& \cellcolor{HIT-L-table}58.3 & \cellcolor{HIT-L-table}23 
& \cellcolor{HIT-L-table}58.3 & \cellcolor{HIT-L-table}47 
& \cellcolor{HIT-L-table}58.3 & \cellcolor{HIT-L-table}21 
& \cellcolor{HIT-L-table}79.2 & \cellcolor{HIT-L-table}107 
& \cellcolor{HIT-L-table}\textbf{83.3} & \cellcolor{HIT-L-table}\textbf{52}
& \cellcolor{HIT-L-table}\textbf{83.3} & \cellcolor{HIT-L-table}\textbf{63} 
& \cellcolor{HIT-L-table}33.3 & \cellcolor{HIT-L-table}15 
& \cellcolor{HIT-L-table}66.7 & \cellcolor{HIT-L-table}106
& \cellcolor{HIT-L-table}33.3 & \cellcolor{HIT-L-table}34
& \cellcolor{HIT-L-table}70.8 & \cellcolor{HIT-L-table}72
& \cellcolor{HIT-L-table}8.33 & \cellcolor{HIT-L-table}7 
\\ \midrule
\textcolor{MXT-Scratch-S}{\textbf{MXT}} & \textcolor{MXT-Scratch-S}{\textbf{N}}& \textcolor{MXT-Scratch-S}{\textbf{smaller}} 
& \cellcolor{MXT-Scratch-S-table}70.8 & \cellcolor{MXT-Scratch-S-table}56
& \cellcolor{MXT-Scratch-S-table}33.3 & \cellcolor{MXT-Scratch-S-table}20 
& \cellcolor{MXT-Scratch-S-table}66.7 & \cellcolor{MXT-Scratch-S-table}54 
& \cellcolor{MXT-Scratch-S-table}41.7 & \cellcolor{MXT-Scratch-S-table}15 
& \cellcolor{MXT-Scratch-S-table}\underline{87.5} & \cellcolor{MXT-Scratch-S-table}109
& \cellcolor{MXT-Scratch-S-table}16.7 & \cellcolor{MXT-Scratch-S-table}10
& \cellcolor{MXT-Scratch-S-table}66.7 & \cellcolor{MXT-Scratch-S-table}52
& \cellcolor{MXT-Scratch-S-table}33.3 & \cellcolor{MXT-Scratch-S-table}14  
& \cellcolor{MXT-Scratch-S-table}62.5 & \cellcolor{MXT-Scratch-S-table}105
& \cellcolor{MXT-Scratch-S-table}16.7 & \cellcolor{MXT-Scratch-S-table}30
& \cellcolor{MXT-Scratch-S-table}75.0 & \cellcolor{MXT-Scratch-S-table}75
& \cellcolor{MXT-Scratch-S-table}33.3 & \cellcolor{MXT-Scratch-S-table}24 
\\
\textcolor{MXT-Scratch-L}{\textbf{MXT}} & \textcolor{MXT-Scratch-L}{\textbf{N}}& \textcolor{MXT-Scratch-L}{\textbf{larger}} 
& \cellcolor{MXT-Scratch-L-table}87.5 & \cellcolor{MXT-Scratch-L-table}\textbf{67} 
& \cellcolor{MXT-Scratch-L-table}\underline{83.3} & \cellcolor{MXT-Scratch-L-table}\underline{31} 
& \cellcolor{MXT-Scratch-L-table}70.8 & \cellcolor{MXT-Scratch-L-table}53
& \cellcolor{MXT-Scratch-L-table}41.7 & \cellcolor{MXT-Scratch-L-table}16 
& \cellcolor{MXT-Scratch-L-table}83.3 & \cellcolor{MXT-Scratch-L-table}107
& \cellcolor{MXT-Scratch-L-table}50.0 & \cellcolor{MXT-Scratch-L-table}37
& \cellcolor{MXT-Scratch-L-table}75.0 & \cellcolor{MXT-Scratch-L-table}60
& \cellcolor{MXT-Scratch-L-table}\underline{58.3} & \cellcolor{MXT-Scratch-L-table}23
& \cellcolor{MXT-Scratch-L-table}62.5 & \cellcolor{MXT-Scratch-L-table}98
& \cellcolor{MXT-Scratch-L-table}\underline{41.7} & \cellcolor{MXT-Scratch-L-table}\underline{38}
& \cellcolor{MXT-Scratch-L-table}79.2 & \cellcolor{MXT-Scratch-L-table}76
& \cellcolor{MXT-Scratch-L-table}33.3 & \cellcolor{MXT-Scratch-L-table}22
\\
\textcolor{MXT-Pretrained-S}{\textbf{MXT}} & \textcolor{MXT-Pretrained-S}{\textbf{Y}}& \textcolor{MXT-Pretrained-S}{\textbf{smaller}}
& \cellcolor{MXT-Pretrained-S-table}\underline{91.7} & \cellcolor{MXT-Pretrained-S-table}66 
& \cellcolor{MXT-Pretrained-S-table}\underline{83.3} & \cellcolor{MXT-Pretrained-S-table}30 
& \cellcolor{MXT-Pretrained-S-table}\underline{83.3} & \cellcolor{MXT-Pretrained-S-table}\underline{62} 
& \cellcolor{MXT-Pretrained-S-table}\underline{83.3} & \cellcolor{MXT-Pretrained-S-table}\underline{31} 
& \cellcolor{MXT-Pretrained-S-table}83.3 & \cellcolor{MXT-Pretrained-S-table}103 
& \cellcolor{MXT-Pretrained-S-table}75.0 & \cellcolor{MXT-Pretrained-S-table}47  
& \cellcolor{MXT-Pretrained-S-table}79.2 & \cellcolor{MXT-Pretrained-S-table}61 
& \cellcolor{MXT-Pretrained-S-table}\underline{58.3} & \cellcolor{MXT-Pretrained-S-table}24 
& \cellcolor{MXT-Pretrained-S-table}\textbf{87.5} & \cellcolor{MXT-Pretrained-S-table}\textbf{129}
& \cellcolor{MXT-Pretrained-S-table}25.0 & \cellcolor{MXT-Pretrained-S-table}35
& \cellcolor{MXT-Pretrained-S-table}\underline{83.3} & \cellcolor{MXT-Pretrained-S-table}\underline{83} 
& \cellcolor{MXT-Pretrained-S-table}\underline{58.3} & \cellcolor{MXT-Pretrained-S-table}\underline{33} 
\\
\textcolor{MXT-Pretrained-L}{\textbf{MXT}} & \textcolor{MXT-Pretrained-L}{\textbf{Y}}& \textcolor{MXT-Pretrained-L}{\textbf{larger}}
& \cellcolor{MXT-Pretrained-L-table}\textbf{95.8} & \cellcolor{MXT-Pretrained-L-table}\textbf{67} 
& \cellcolor{MXT-Pretrained-L-table}\textbf{91.7} & \cellcolor{MXT-Pretrained-L-table}\textbf{34} 
& \cellcolor{MXT-Pretrained-L-table}\textbf{91.7} & \cellcolor{MXT-Pretrained-L-table}\textbf{67} 
& \cellcolor{MXT-Pretrained-L-table}\textbf{100} & \cellcolor{MXT-Pretrained-L-table}\textbf{36}
& \cellcolor{MXT-Pretrained-L-table}\textbf{95.8} & \cellcolor{MXT-Pretrained-L-table}\textbf{116}
& \cellcolor{MXT-Pretrained-L-table}\textbf{83.3} & \cellcolor{MXT-Pretrained-L-table}\textbf{52}
& \cellcolor{MXT-Pretrained-L-table}\textbf{83.3} & \cellcolor{MXT-Pretrained-L-table}\textbf{63}
& \cellcolor{MXT-Pretrained-L-table}\textbf{75.0} & \cellcolor{MXT-Pretrained-L-table}\textbf{29}
& \cellcolor{MXT-Pretrained-L-table}\textbf{87.5} & \cellcolor{MXT-Pretrained-L-table}\textbf{129}
& \cellcolor{MXT-Pretrained-L-table}\textbf{66.7} & \cellcolor{MXT-Pretrained-L-table}\textbf{52}
& \cellcolor{MXT-Pretrained-L-table}\textbf{91.7} & \cellcolor{MXT-Pretrained-L-table}\textbf{88}
& \cellcolor{MXT-Pretrained-L-table}\textbf{83.3} & \cellcolor{MXT-Pretrained-L-table}\textbf{42}
\\
\bottomrule
\end{tabular}
\end{adjustbox}
\end{center}
\begin{flushleft}
* Number of trajectories: TC-Uni smaller=20, larger=40; TC-Bi smaller=30, larger=60; SO-Uni smaller=40, larger=80; SO-Bi smaller=40, larger=80; Scoop-Uni smaller=30, larger=60; Pour-Bi smaller=30, larger=60.
\end{flushleft}
\label{table:main}
\end{table*}

%% file: arxiv_contents/limitations.tex
\section{Limitations}
\label{sec:limitations}



While our system introduces a novel and effective approach for cross-embodiment learning and data collection in quadrupedal manipulation, it has several limitations. 
First, although the teleoperation interface enables whole-body control and flexible data collection, it still requires a learning curve for operators. In particular, controlling the robot’s torso via head motions may feel unintuitive and demand some practice to achieve high precision. 
Second, despite the modular design and cross-embodiment capability of our architecture, this work focuses solely on the LocoMan platform. We have not yet validated the generality of our system across a broader range of robot embodiments, such as table-top or mobile arms and humanoid robots. 
Lastly, while we have demonstrated effective human-to-robot transfer, we rely on a pretraining-finetuning pipeline for single robotic tasks, and future work could explore co-training with human data in multi-task settings. Extending Human2LocoMan to diverse robots, larger-scale datasets, and more learning paradigms remains promising future directions.

%% file: arxiv_contents/conclusion.tex
\section{Conclusion}
\label{sec:conclusion}


We present \oursystem, a unified framework for efficient data collection and cross-embodiment learning, enabling versatile quadrupedal manipulation skills on the open-source LocoMan platform. 
Our system integrates XR-based teleoperation with aligned human and robot data collection, supporting a shared observation-action representation across embodiments. To leverage this data effectively, we introduce the Modularized Cross-embodiment Transformer, a modular policy architecture designed for scalable pretraining and fine-tuning across different embodiments and manipulation modes.
Extensive real-world experiments across six challenging household tasks demonstrate that \oursystem achieves strong performance, robust generalization to out-of-distribution settings, and high data efficiency, attaining high success rates with only a small amount of robot data. Comparisons against strong baselines highlight the effectiveness of our modular design and the benefits of cross-embodiment pretraining using human data collected through our system.

%% file: contents/appendix.tex
\appendix
\label{sec:appendix}

\subsection{Implementation and Training details of MXT}
\label{sec:MXT para}

\parab{Training Details.} We list the training optimizer and the Transformer trunk hyperparamters in Table \ref{Tab:MXT}. These hyperparameters are kept the same for all our experiments.

\begin{table}[htp]
\caption{MXT trunk and training hyperparameters}
\label{Tab:MXT}
\centering
\begin{tabular}{@{}lc@{}}
\toprule
Hyperparameters                & Value                          \\ \midrule
optimizer                      & AdamW                          \\
\multirow{2}{*}{learning rate} & 5e-5 (finetuning/from scratch) \\
                               & 1e-4 (pretraining)              \\
scheduler                      & constant                       \\
weight decay                   & 1e-4                           \\ \midrule
trunk encoder layers                 & 4                              \\
trunk decoder layers                 & 4                              \\
hidden dim.                     & 128                            \\
Transformer feedforward dim.                & 256                            \\
\#attention heads              & 16                             \\ \bottomrule
\end{tabular}
\end{table}

\parab{Cross-Attention in Tokenizers and Detokenizers.}
In the tokenizers of MXT, we employ a simple cross-attention mechanism to convert input features of arbitrary length into a fixed set of tokens. Each attention layer in the tokenizers uses a hidden dimension of 128, with 4 attention heads, each having a head dimension of 32, and a dropout rate of 0.1.
Other hyperparameters of each tokenizer are shown in Table \ref{Tab:MXT tokenizer}.

Similarly, we use cross-attention in the detokenizers to decode action modalities from a fixed number of output Transformer tokens. The attention layer is configured with 4 attention heads, each with a head dimension of 16, and a dropout rate of 0.1.
Other hyperparameters of each detokenizer are shown in Table \ref{Tab:MXT detokenizer}

\begin{table}[htp]
\caption{MXT tokenizer hyperparameters}
\label{Tab:MXT tokenizer}
\centering
\begin{tabular}{@{}lcccc@{}}
\toprule
Modality       & Input dimensions & \#tokens & MLP widths                                \\ \midrule
main image       & (3, 480 1280)    & 16       & \multirow{2}{*}{N/A}            \\ 
wrist image      & (3, 480, 640)    & 8        &                                                             \\ \midrule
body pose        & (6,)             & 4        & \multirow{4}{*}{{[}128, 128{]}}        \\
EEF pose         & (12,)            & 4        &                                                           \\
EEF-to-body pose & (12,)            & 4        &                                                           \\
gripper angles    & (2,)             & 4        &                                                            \\ \bottomrule
\end{tabular}
\end{table}

\begin{table}[htp]
\caption{MXT detokenizer hyperparameters}
\label{Tab:MXT detokenizer}
\centering
\begin{tabular}{@{}lcc@{}}
\toprule
Modalities    & Output dimensions & \#tokens \\ \midrule
body pose     & (6,)              & 6        \\
EEF pose      & (12,)             & 6        \\
gripper angle & (2,)              & 6        \\ \bottomrule
\end{tabular}
\end{table}

\parab{Masks for aligning embodiment modalities.}
As previously mentioned, masks are required to exclude redundant dimensions or modalities that are absent in certain embodiments. Below, we provide a more detailed description of our mask implementation.

\textit{a) Masks on images.}
We recognize that some image views may not be available across all embodiments. In the ecperiments of this paper, we assume at most two camera views (or image modalities): a main camera and a wrist camera. However, the framework can be easily extended to support any number of camera views. When a particular view is unavailable, we indicate this in the Transformer's trunk mask and insert dummy tokens at the corresponding positions. This ensures that tokens associated with the missing image modality are not attended to.

\textit{b) Masks on proprioceptive states.}
In some cases, certain dimensions of the proprioceptive state may be irrelevant for a specific embodiment. For example, in unimanual manipulation modes for both the human and LocoMan, the proprioceptive states of the inactive end-effector are not used. When only part of a proprioceptive modality contains redundant dimensions, we apply zero padding to those dimensions and encode them through the tokenizer as usual. Unlike masked image modalities, this does not affect the Transformer's trunk mask. However, when an entire proprioception modality should be disregarded, we handle it similarly to image modalities by applying the Transformer mask accordingly.


\parab{Data Normalization.} 
For both human and LocoMan data, we apply normalization to observations and action labels. For non-image data, we compute the per-dimension mean and standard deviation from the dataset, and normalize using the standard formula:
\[
\bar{x}_t = \frac{x_t - \text{mean}}{\text{std}}.
\]
For image data, we adopt the standard ImageNet statistics for the RGB channels to normalize each image in the dataset using the same formula, with 
\(\text{mean} = [0.485, 0.456, 0.406]\) and 
\(\text{std} = [0.229, 0.224, 0.225]\).

\parab{Dropout in Pretraining.} 
We find that increasing the dropout rate in the Transformer trunk improves finetuning performance for MXT. In practice, setting the pretraining dropout rate to 0.5 for the scooping task and 0.4 for all other tasks yields consistently good results. When training with LocoMan data, whether training from scratch or during finetuning, the Transformer trunk dropout rate is reverted to 0.1.

\begin{table}[htp]
\caption{HIT hyperparameters}
\label{Tab:HIT}
\centering
\begin{tabular}{@{}lc@{}}
\toprule
Hyperparameters     & Value    \\ \midrule
optimizer           & AdamW    \\
learning rate       & 2e-5     \\
scheduler           & constant \\
weight decay        & 1e-4     \\
encoder layers      & 4        \\
decoder layers      & 4        \\
hidden dim          & 128      \\
\#attention heads   & 8        \\
feature loss weight & 0.001    \\
image backbone      & ResNet18 \\ \bottomrule
\end{tabular}
\end{table}

\begin{table}[htp]
\caption{HPT hyperparameters}
\label{Tab:HPT}
\centering
\begin{tabular}{@{}lc@{}}
\toprule
Hyperparameters                       & Value                          \\ \midrule
optimizer                             & AdamW                          \\
\multirow{2}{*}{learning rate}        & 5e-5 (finetuning/from scratch) \\
                                      & 1e-4 (pretraining)              \\
scheduler                             & constant                       \\
weight decay                          & 1e-4                           \\ \midrule
\textbf{trunk}                           &                                \\
\multicolumn{1}{r}{\#Transformer blocks}                  & 16                             \\
\multicolumn{1}{r}{hidden dim}                            & 128                            \\
\multicolumn{1}{r}{feedforward dim}                       & 256                            \\
\multicolumn{1}{r}{\#attention heads}                     & 8                              \\ \midrule
\textbf{action head}                           &                                \\
\multicolumn{1}{r}{\#attention heads} & 8                              \\
\multicolumn{1}{r}{head dim}          & 64                             \\
\multicolumn{1}{r}{dropout}           & 0.1                            \\
\multicolumn{1}{r}{output dim}        & 20                             \\ \midrule
\textbf{image stem}                            &                                \\
\multicolumn{1}{r}{encoder}           & ResNet18                       \\
\multicolumn{1}{r}{MLP widths}        & {[}128{]}                      \\
\multicolumn{1}{r}{\#tokens}          & 16                             \\ 
\textbf{state stem}                            &                                \\
\multicolumn{1}{r}{MLP widths}        & {[}128{]}                      \\
\multicolumn{1}{r}{\#tokens}          & 16                             \\ \bottomrule
\end{tabular}
\end{table}


\subsection{Implementation details of baselines}
\label{sec:baseline para}
\parab{HIT.} Our implementation of Humanoid Imitation Transformer \cite{fu2024humanplus} is based on the released codebase, with only minor modifications to accommodate our data format. The hyperparameters used for training are summarized in Table \ref{Tab:HIT}.

\parab{HPT.} 
We follow the original implementation of HPT~\cite{wang2024scaling}, with the main exception of modifying the data normalization method to align with the approach used in other frameworks. This ensures a fair comparison of validation loss. The hyperparameters used for training HPT are summarized in Table~\ref{Tab:HPT}.


\subsection{Global task-specific training parameters}
\label{sec:global para}
We select a specific set of training parameters for each task and keep these settings consistent across all methods, as summarized in Table~\ref{Tab:GlobalTrain}.

\begin{table}[htp]
\caption{Global training parameters for each task}
\label{Tab:GlobalTrain}
\centering
\begin{tabular}{@{}lcccc@{}}
\toprule
Task                                      & Mode     & Batch Size & Training Steps & Chunk Size \\ \midrule
\multirow{2}{*}{Toy Collection}           & Unimanual   & 16         & 60000          & 60         \\
                                          & Bimanual & 16         & 60000          & 60         \\ \midrule
\multirow{2}{*}{Shoe Organization} & Unimanual   & 24         & 80000          & 180        \\
                                          & Bimanual & 24         & 100000         & 120        \\ \midrule
Scooping                                 & Unimanual & 24         & 100000         & 120        \\ \midrule
Pouring                                   & Bimanual & 24         & 80000          & 180        \\ \bottomrule
\end{tabular}
\end{table}